\pdfoutput=1

\documentclass[11pt,x11names]{article}

\usepackage[preprint]{acl}
\usepackage{amsmath, amssymb}
\usepackage{enumitem}
\usepackage{times}
\usepackage{latexsym}
\usepackage{bbm}
\usepackage{booktabs}
\usepackage{multirow}
\usepackage{makecell}
\usepackage{longtable, array}

\usepackage[T1]{fontenc}
\usepackage[ruled,vlined]{algorithm2e}
\usepackage{array}
\usepackage[utf8]{inputenc}
\usepackage{microtype}
\usepackage{hyperref}  
\usepackage{footnote}  
\usepackage{inconsolata}
\usepackage{xcolor}
\usepackage{soul}
\usepackage{graphicx}
\usepackage{float}
\definecolor{softblue}{RGB}{100, 149, 237}
\usepackage{hyperref}  
\hypersetup{
    colorlinks,
    linkcolor=SeaGreen4,
    anchorcolor=blue,
    citecolor=RoyalBlue3
}
\AtBeginDocument{%
  \setlength{\abovedisplayskip}{7pt}
  \setlength{\belowdisplayskip}{7pt}
  \setlength{\abovedisplayshortskip}{7pt}
  \setlength{\belowdisplayshortskip}{7pt}
}
\definecolor{lightgreen}{RGB}{200, 255, 200}  
\definecolor{deepgreen}{RGB}{122, 255, 122}       

\newcommand{\lighthl}[1]{\sethlcolor{lightgreen}\hl{#1}}
\newcommand{\deephl}[1]{\sethlcolor{deepgreen}\hl{#1}}

\title{SAIF: A Sparse Autoencoder Framework for Interpreting and Steering Instruction Following of Language Models}

\author{Zirui He\textsuperscript{1,*}, Haiyan Zhao\textsuperscript{1,*}, Yiran Qiao\textsuperscript{2}, Fan Yang\textsuperscript{3}, \\
\textbf{Ali Payani\textsuperscript{4}}, \textbf{Jing Ma\textsuperscript{2}}, \textbf{Mengnan Du\textsuperscript{1}}\\
\textsuperscript{1}NJIT \,
\textsuperscript{2}Case Western Reserve University \,
\textsuperscript{3}Wake Forest University \,
\textsuperscript{4}Cisco\\
\textsuperscript{*}Equal contribution\\
\small\texttt{\{zh296,hz54,mengnan.du\}@njit.edu}, \small\texttt{\{yxq350,jxm1384\}@case.edu},
\small\texttt{yangfan@wfu.edu}, \small\texttt{apayani@cisco.com}
}

\begin{document}

\maketitle
\begin{abstract}
The ability of large language models (LLMs) to follow instructions is crucial for their practical applications, yet the underlying mechanisms remain poorly understood. This paper presents a novel framework that leverages sparse autoencoders (SAE) to interpret how instruction following works in these models. We demonstrate how the features we identify can effectively steer model outputs to align with given instructions. Through analysis of SAE latent activations, we identify specific latents responsible for instruction following behavior. Our findings reveal that instruction following capabilities are encoded by a distinct set of instruction-relevant SAE latents. These latents both show semantic proximity to relevant instructions and demonstrate causal effects on model behavior. Our research highlights several crucial factors for achieving effective steering performance: precise feature identification, the role of final layer, and optimal instruction positioning. Additionally, we demonstrate that our methodology scales effectively across SAEs and LLMs of varying sizes.


\end{abstract}

\section{Introduction}

Large language models (LLMs) have demonstrated remarkable capabilities in following instructions, enabling alignment between model outputs and user objectives. These capabilities are typically gained through instruction tuning methods~\cite{ouyang2022training, wei2022finetuned}, including extensive training data and computationally intensive fine-tuning processes. While these approaches effectively control model behavior, the underlying mechanisms by which models process and respond to instructions remain poorly understood. In-depth mechanistic investigations are essential for improving our ability to control models and enhance their instruction-following capability.

Prior research has attempted to understand instructions following from two perspectives: 1) prompting-based; 2) activation-space-based. Among prompting-based studies, the importance of instruction positions has been thoroughly studied~\cite{liu2023instruction,ma2024getting}. For activation-based studies,~\citet{stolfo2024improving} propose to manipulate model following instructions with representation vector in residual stream. However, both methods ultimately fail to explain the inner workings of how LLMs follow instructions in a fine-grained manner, i.e. the concept level. Specifically, prompting-based approaches provide insights into better prompt formulation strategies to improve instruction following, while activation-space-based methods provide a possible way to implement steering with instruction following rather than explaining how it works.

\begin{figure*}[tb]
    \centering
    \includegraphics[width=\textwidth]{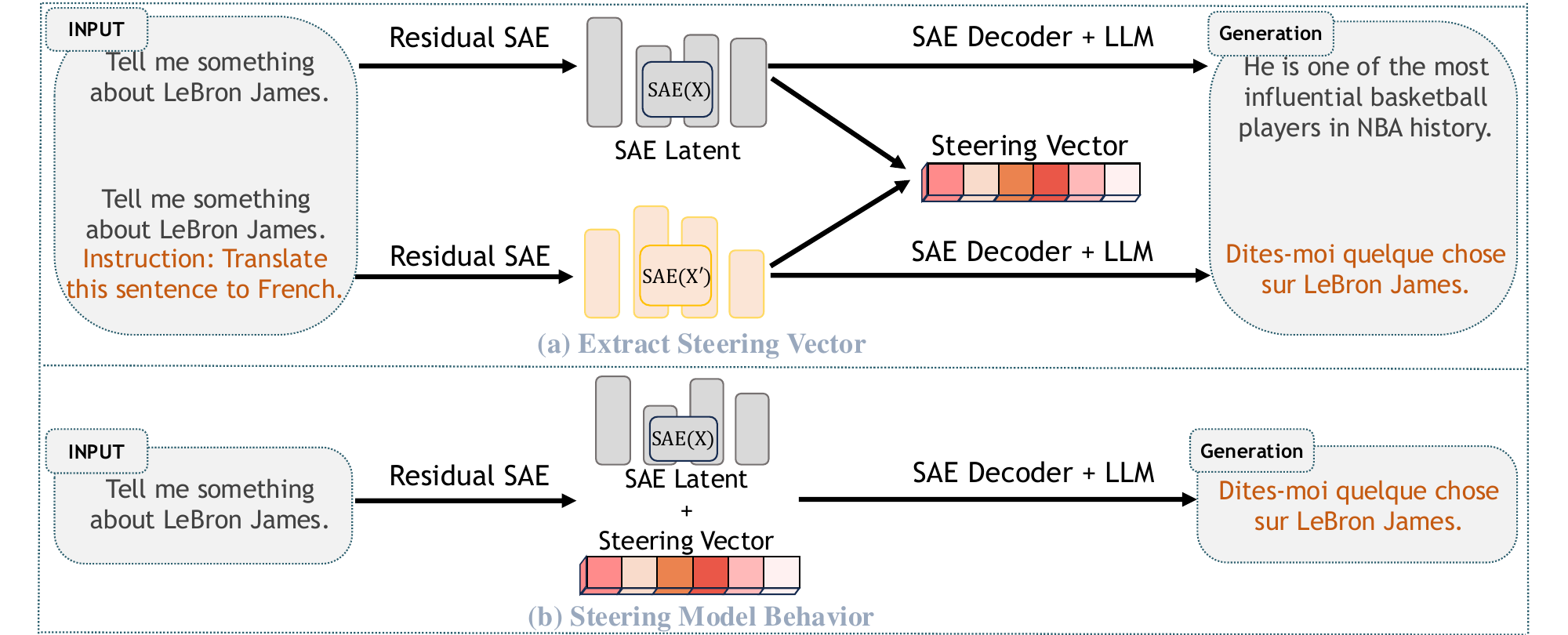}
    \caption{The proposed SAIF framework. The model computes steering vectors from SAE latent differences to guide outputs according to instructions. (a) Extract steering vector. (b) Apply steering for controlled output.}
    \label{fig:overall-pipeline}
\end{figure*}

In this paper, we propose a novel framework \textbf{SAIF} (\underline{S}parse \underline{A}utoencoder steering for \underline{I}nstruction \underline{F}ollowing) to understand working mechanisms of instruction following at the concept level through the lens of sparse autoencoders (SAEs). First, we develop a robust method to sample instruction-relevant features. Then, we select influential features using designed metrics and further compute steering vectors (see Figure~\ref{fig:overall-pipeline}a). Furthermore, we measure the effectiveness of these steering vectors through steering tasks (see Figure~\ref{fig:overall-pipeline}b). Additionally, we examine the extracted features using Neuronpedia~\cite{neuronpedia} to illustrate how semantically relevant the activating text of features is to instructions. We also measure steering performance to demonstrate the effectiveness of extracted features. Through these tools, we gain some intriguing insights regarding the importance of the feature number used in representing instructions, the role of the last layer, the impact of instruction position and model scale.
\noindent
Our main contributions in this work can be summarized as follows:
\vspace{-3pt}
\begin{itemize}[leftmargin=*]\setlength\itemsep{-0.3em}
\item We propose \textbf{SAIF}, a framework that interprets instruction following in LLMs at a fine-grained conceptual level. Our analysis reveals how models internally encode and process instructions through interpretable latent features in their representation space.
\item We demonstrate that instructions cannot be adequately represented by a single concept in SAEs, but rather comprise multiple high-level concepts. Effective instruction steering requires a set of instruction-relevant features, which our method precisely identifies.
\item We reveal the critical role of the last layer in SAE-based activation steering. Moreover, the effectiveness of our framework has been demonstrated across instruction types and model scales.
\end{itemize}

\section{Preliminaries}\label{sec:pre}

\paragraph{Sparse Autoencoders (SAEs).} Dictionary learning enables disentangling representations into a set of concepts~\cite{olshausen1997sparse,bricken2023towards}. SAEs are employed to decompose hidden representations into a high-dimension space and then reconstruct the hidden representations. Specifically, the input of SAEs is the hidden representation from a model's residual stream denoted as $\boldsymbol{z} \in \mathbb{R}^d$ and the reconstructed output is denoted as $\text{SAE}(\boldsymbol{z}) \in \mathbb{R}^d$, we obtain that $\boldsymbol{z} = \text{SAE}(\boldsymbol{z}) + \epsilon$ where $\epsilon$ is the error. In our paper, we focus on layerwise SAEs trained with an encoder $\boldsymbol{W}_\text{enc} \in \mathbb{R}^{d \times m}$ followed by the non-linear activation function, and a decoder $\boldsymbol{W}_\text{dec} \in \mathbb{R}^{m \times d}$~\cite{he2024llama}. The definition of SAEs is:
\begin{align}
&a(\boldsymbol{z})=\sigma\left(\boldsymbol{z} \boldsymbol{W}_{\mathrm{enc}}\right.+\left.\boldsymbol{b}_{\mathrm{enc}}\right), \label{eq:1} \\
&\operatorname{SAE}(\boldsymbol{z})=a(\boldsymbol{z}) \boldsymbol{W}_{\mathrm{dec}}\ +\ \boldsymbol{b}_{\mathrm{dec}}, \label{eq:2}
\end{align}
where $\boldsymbol{b}_\text{enc} \in \mathbb{R}^m$ and $\boldsymbol{b}_\text{dec} \in \mathbb{R}^d$ are the bias terms. The decomposed high-dimension latent activations $a(\boldsymbol{z})$ have dimension $m$ and $m \gg d$, which is a highly sparse vector. 
Note that different SAEs use different non-linear activation function $\sigma$. For example, Llama Scope~\cite{he2024llama} adopts TopK-ReLU, while Gemma Scope~\cite{lieberum2024gemma} uses JumpReLU~\cite{rajamanoharan2024jumpingaheadimprovingreconstruction}.

\paragraph{Steering with SAE Latents.} Following Eq.~\eqref{eq:2}, the reconstructed SAE outputs are a linear combination of \textit{SAE latents}, which represent the row vectors of SAE decoder $\boldsymbol{W}_\text{dec}$. The weight of $j$-th SAE latent is $a(\boldsymbol{z})_j$. Typically, a prominent dimension $j \in \{1, \cdots, m\}$ is chosen, and its decoder latent vector $\boldsymbol{d}_j$ is scaled with a factor $\alpha$ and then added to the SAE outputs~\cite{ferrando2025do}. The computation is as follows:
\begin{equation}
\boldsymbol{z}^{\text {new }} \leftarrow \boldsymbol{z}+\alpha \boldsymbol{d}_j.\label{eq:steer}
\end{equation}
This modified representation $z^{\text{new}}$ can then be fed back into the model's residual stream to steer the model's behavior during generation. 

\section{Proposed Method}

In this section, we introduce the \textbf{SAIF}, a framework for analyzing and steering instruction following in LLMs. First, 
we introduce linguistic variations to construct diverse instruction sentences and related datasets, which are further used to compute SAE latent activations.
Second, we develop a two-stage process for computing steering vectors that quantifies the sensitivity of features to instruction presence. Finally, we investigate how these identified features can be leveraged for steering model behavior, demonstrating a technique for enhancing instruction following while preserving output coherence (see Figure~\ref{fig:overall-pipeline}).



\subsection{Format Instruction Feature}\label{sec:for-ins-fea}

To identify instruction-relevant features given an instruction, we construct a dataset $\mathcal{D}$ with $N$ positive-negative sample pairs. For example, we focus on an instruction \colorbox{LavenderBlush1}{{\texttt{Translate the sentence to}}} \colorbox{LavenderBlush1}{{\texttt{ French}}}. In a sample pair, the positive sample refers to a prefix prompt followed by the instruction, while the negative sample refers to the prefix prompt without the instruction sentence. 

The \textit{difference-in-means}~\cite{rimsky2024steering} is a typical approach to derive concept vectors. It computes the activation differences between each sample pair over the last token, and then averages over all pairs of activation difference vectors. However, directly applying this pipeline to instruction following presents a significant challenge. When a single instruction sentence is used repeatedly to generate samples, the model tends to encode the specific semantic meaning of that instruction rather than learning a general-purpose vector that can reliably execute the intended operation (See Appendix~\ref{app:instruct-example}). Specifically, the derived vector can barely operate the same instruction if we rephrase the instruction in a linguistically different but semantically similar manner. To resolve this challenge, we propose to introduce linguistic variations to extract instruction functions.

We formulate instruction sentences for a given instruction through different strategies. These variations include syntactic reformulations (e.g., imperative to interrogative form, task-oriented to process-based description) and cross-lingual translations (e.g., English, Chinese, German). In this way, we generated six diverse instruction sentences comprehensively capturing key features of an instruction. The instruction design used in our paper is shown in Appendix~\ref{app:a}. 



For each instruction variant, we extract samples' residual stream representation and compute the corresponding SAE latent activations. While diverse linguistic information are contained, the latent features specifically corresponding to the core instructional concept should maintain relatively consistent activation levels across all variants. These dimensions with consistent activation patterns will be further used to construct instruction vectors.



\subsection{Steering Vector Computation}\label{sec:steer-vec-compute}
Based on SAE latent activations computed in Section~\ref{sec:for-ins-fea}, we develop a two-step process for computing steering vectors. The first step identifies features that consistently respond to a given instruction, while the second step quantifies their sensitivity.

Given $N$ input samples and a target instruction type (e.g., translation), we first obtain both positive samples (with instruction) and negative samples (without instruction) for each input. For each sample pair $i$ and feature $j$, we compute the activation state change:
\begin{equation}
\Delta h_{i,j} = \mathbbm{1}(h_{i,j}^{\text{w}} > 0) - \mathbbm{1}(h_{i,j}^{\text{w/o}} > 0), \label{eq:response}
\end{equation}
where $h_{i,j}^{\text{w}}$ and $h_{i,j}^{\text{w/o}}$ represent the SAE latent activation values with and without instruction respectively, and $\mathbbm{1}(\cdot)$ is the indicator function. $\Delta h_{i,j}$ captures whether feature $j$ becomes activated in response to the instruction for sample $i$.
We then compute a sensitivity score $C_j$ for each feature:
\begin{equation}
C_j = \frac{1}{N}\sum_{i=1}^N \mathbbm{1}(\Delta h_{i,j} > 0). \label{eq:sensitivity}
\end{equation}
The score represents the proportion of samples whose feature $j$ becomes activated in response to instructions. Features with higher scores are more consistently responsive to instruction presence.
By sorting these sensitivity scores in a descending order, we select the top-$k$ responsive features. These selected features form the instruction-relevant feature set $\boldsymbol{V} = \{\boldsymbol{W}_{\text{dec}, j}| \text{rank}(C_j) \leq k\}$ where $\boldsymbol{W}_{\text{dec}, j} = \boldsymbol{W}_{\text{dec}}[j,:]$ denotes the $j$-th SAE latent. These features will be used for further constructing steering vectors.

\begin{algorithm}[t!]\small
\DontPrintSemicolon
\KwIn{Input text $x$; Target instruction type (e.g., translation, summarization)}

\textbf{Stage 1: Format Instruction Feature}\;
Generate diverse instruction variants\;
Construct dataset $\mathcal{D}$ with $N$ positive/negative pairs\;

\textbf{Stage 2: Compute Steering Vector}\;
\For{each sample pair $i$ and feature $j$}{
    Compute activation state change:\;
    $\Delta h_{i,j} = \mathbbm{1}(h_{i,j}^w > 0) - \mathbbm{1}(h_{i,j}^{w/o} > 0)$\;
    Calculate sensitivity score:\;
    $C_j = \frac{1}{N}\sum_{i=1}^N \mathbbm{1}(\Delta h_{i,j} > 0)$\;
}
Sort features by sensitivity scores $C_j$\;
Select top-$k$ features as instruction-relevant set $\boldsymbol{V}$\;

\textbf{Stage 3: Steering Procedure}\;
Obtain residual stream representation $z$ of input $x$\;
\For{each feature $i \in \boldsymbol{V}$}{
    Compute activation strength:
    $\alpha_i = \mu_i + \beta s_i$\;
    where $\mu_i$ is mean activation, $s_i$ is std deviation\;
}
Apply steering:
$\boldsymbol{z}^{new} = \boldsymbol{z} + \sum_{i=1}^k \alpha_i \boldsymbol{v}_i$\;

\KwOut{Steered text following the instruction}
\caption{\small The proposed \textbf{SAIF} framework}
\label{alg:SAIF}
\end{algorithm}

\subsection{Steering Procedure}

Different from the classic steering approach defined in Eq.~\eqref{eq:steer}, we hypothesize that instruction following steering requires a set of features to be effective. The individual feature utilized in the classic method focuses on token-level concepts, where individual concepts typically correlate with a few SAE latent activations. As a result, this approach can barely operate instructions. It is partly due to the complexity of sentence-level instructions, which are composed of multiple high-level features represented by a set of SAE latent features. Additionally, SAEs tend to overly split features, which further increases the number of features needed for steering~\cite{ferrando2025do}. Thus, we propose to determine how to steer with a set of vectors.

Building on top of the feature set $\boldsymbol{V}$ derived in Section~\ref{sec:steer-vec-compute}, we employ the set of features to steer residual stream representation of a certain input at layer $l$. Our steering is implemented as below:
\begin{equation}
\boldsymbol{z}^\text{new} = \boldsymbol{z} + \sum_{i=1}^k \alpha_i \boldsymbol{v}_i,
\label{equ:steering}
\end{equation}
where $\boldsymbol{z}$ represents the residual stream representation of the input over the last token, and $\alpha_i$ denotes the steering strength of feature $i$. Here, $\boldsymbol{v}_i$ represents a certain instruction-relevant feature in $\boldsymbol{V}$.

As the strength of each selected feature is crucial to steering performance, we further compute the strength of each feature by employing statistical measurements of feature activation values to make it more robust and reliable. The activation strength for feature $i$ is calculated as:
\begin{equation}
\alpha_i = \mu_i + \beta s_i,
\label{equ:st}
\end{equation}
where $\mu_i$ is the mean activation value of feature $i$ observed in instruction-following examples, $s_i$ is the standard deviation of these activation values, and $\beta$ is a hyperparameter to scale $s_i$ meanwhile controlling the strength value.


\begin{figure*}[tb]
    \centering
    \includegraphics[width=\textwidth]{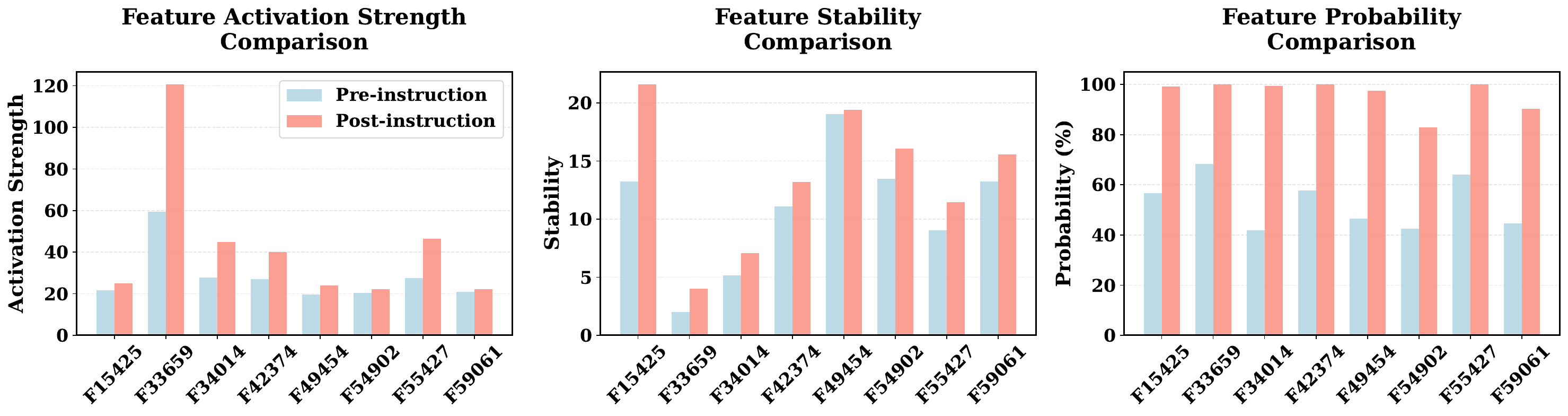}
    \caption{Comparison of feature activation patterns between pre-instruction and post-instruction conditions across different SAE latent dimensions. The plots show three key metrics: activation strength (left), feature stability (middle), and activation probability (right) for eight identified instruction-relevant features.}
    \label{fig:pre-post-instruction}
\end{figure*}

\section{Experiments}
In this section, we conduct experiments to evaluate the effectiveness of \textbf{SAIF}  by answering the following research questions (RQs):
\vspace{-5pt}
\begin{itemize}[leftmargin=*]\setlength\itemsep{-0.3em}
   \item RQ1: How interpretable are the features extracted using SAEs, and do they correspond to instruction-related concepts? (Section 4.2) 
   
   \item RQ2: Can the proposed \textbf{SAIF} framework effectively control model behavior? (Section 4.3)
   
   \item RQ3: What role does the final Transformer layer play in the instruction following?
   (Section 4.4)
   
   \item RQ4: How does instruction positioning affect the effectiveness of instruction following and feature activation patterns? (Section 4.5)
\end{itemize}

\subsection{Experimental Setup}

\vspace{3pt}
\noindent\textbf{Datasets and Models.}\,
Our experiments are conducted with multiple language models including Gemma-2-2b, Gemma-2-9b~\cite{team2024gemma} and Llama3.1-8b. The Cross-lingual Natural Language Inference (XNLI) dataset \cite{conneau2018xnli} is used to construct input samples. It encompasses diverse languages (including English, French, Spanish, German, Greek, Bulgarian, Russian, Turkish, Arabic, Vietnamese, Thai, Chinese, Hindi, Swahili and Urdu) and rich syntactic structures (such as active/passive voice alternations, negation patterns, and various clause structures). The diverse linguistic patterns within the dataset are essential in constructing a comprehensive set of samples for an instruction. Moreover, it ensures extracting consistent SAE activations from the residual stream of input samples.

\vspace{3pt}
\noindent\textbf{Instruction Design.}\, 
Following the settings in IFEval~\cite{zhou2023instruction}, we investigate three types of instructions: keyword inclusion, summarization, and translation. For keywords inclusion, we provide models with a keyword (e.g., ``Sunday''), and expect model output incorporating the specified keyword. For formatting, we instruct the model to perform summarization, where the ideal output should be concise, maintain the key information from the original text, and follow a consistent format with a clear topic sentence followed by supporting details. For translation, we direct the model to translate sentences into different languages (English, French, and Chinese), where the ideal model output should accurately perform the requested translation while preserving the original meaning. The complete set of instructions used for each task is provided in Appendix~\ref{app:a}.

\vspace{3pt}
\noindent\textbf{Implementation Details.}\,
We use pre-trained SAEs from Gemma Scope~\cite{lieberum2024gemma} and Llama Scope~\cite{he2024llama}. When constructing input samples for each instruction, we set the number of positive/negative samples $N$ to 800. For SAE latent extraction, we use sparse autoencoders with dimensions of 65K and 131K for \href{https://huggingface.co/google/gemma-2-2b-it}{Gemma-2-2b-it\footnote{https://huggingface.co/google/gemma-2-2b-it}} and \href{https://huggingface.co/google/gemma-2-9b-it}{Gemma-2-9b-it\footnote{https://huggingface.co/google/gemma-2-9b-it}} models respectively. We also use SAE with dimension 32K for \href{https://huggingface.co/meta-llama/Llama-3.1-8B-Instruct}{Llama3.1-8b\footnote{https://huggingface.co/meta-llama/Llama-3.1-8B-Instruct}}. All experiments were run on 1 NVIDIA A100 GPU.
As default settings, for Equation~(\ref{equ:steering}), we fix $k = 15$, meaning that we use the top 15 most responsive SAE features for instruction steering. The strategy to choose the optimal $k$ will be further discussed in Section~\ref{sec:analysis-concept}. For Equation~(\ref{equ:st}), we fix the hyperparameter $\beta=0$, and we discuss the impact of adjusting this hyperparameter on the steering effect in Appendix~\ref{app:C}.

\vspace{3pt}
\noindent\textbf{SAE Latent Activation Metrics.}\, We consider the following three metrics to quantify features' behavior and reliability in instruction processing. Note that we only consider features activated on positive samples but not negative samples.
\begin{itemize}[leftmargin=*]\setlength\itemsep{-0.3em}
    \item \emph{Activation Strength}: The mean activation value is calculated as: $\mu_i = \frac{1}{|A_i|}\sum_{a \in A_i} a$, where $A_i$ is the set of non-zero activation values for feature $i$.
    
    \item \emph{Activation Probability}: The probability of feature $i$ is activated across positive/negative samples: $P_i = \frac{|A_i|}{N}$, where $N$ is the total number of positive/negative samples.
    
    \item \emph{Activation Stability}: The normalized standard deviation value of non-zero activation values: $\Omega_i = 1 / s_i$.
\end{itemize}
A high-quality instruction-relevant feature should ideally exhibit strong activation ($\mu_i$), consistent triggering ($P_i$), and stable behavior ($\Omega_i$) across different formulations of the same instruction.

\vspace{3pt}
\noindent\textbf{Steering Effectiveness Metrics.}\,
We evaluate steering outputs with two metrics: 1) \emph{Strict Accuracy}, which measures the proportion of cases where the model completely follows the instruction, meaning it both understands and produces output exactly as instructed; and 2) \emph{Loose Accuracy}, which measures the proportion of cases where the model partially follows the instruction, meaning it understands the instruction but the output does not fully conform to the requirements. Note that we use GPT-4o-mini to rate the responses, and please refer to the details in Appendix~\ref{sec:gpt-4o-evaluation-details}.

\begin{table*}
\small
\centering
\caption{Maximally activating examples for Feature 15425 in Layer 25 of Gemma2-2b-it when prompted with ``Translate the sentence to French.'' Data sourced from Neuronpedia~\cite{neuronpedia}.}
\begin{tabular}{p{\textwidth}}
\toprule
\multicolumn{1}{c}{Activating Examples with `Translate the sentence to French' (Feature 15425, Layer 25)} \\
\midrule

The Theory of Super conductivity \lighthl{(}195\lighthl{8)} \deephl{(} \lighthl{translated} \deephl{from} \lighthl{Russian:} Consultants Bureau, Inc., New York. \\[0.5em]

Save your game, go back to change the PS 3 system \deephl{language} \lighthl{settings} \deephl{to} \lighthl{English}. \\[0.5em]

We have posted \deephl{a partial} \lighthl{translation} \deephl{of his} \lighthl{speech} \deephl{from} \lighthl{Yiddish} \deephl{to} \lighthl{Hebrew} \deephl{,} \lighthl{which} \deephl{was posted in}... \\[0.5em]

I \lighthl{can} \deephl{speak} \lighthl{English}, \lighthl{but} i'm \deephl{afraid} it \lighthl{may} \deephl{be} \lighthl{worse} \deephl{than your} \lighthl{french}. \\[0.5em]

\bottomrule
\end{tabular}
\label{tab:Feature15425}
\end{table*}

\begin{table*}[t]
    \begin{minipage}{0.55\textwidth}
        \centering
        \caption{Layer25 Experimental Results}
        \scalebox{1}{
        \begin{footnotesize}
        \begin{tabular}{@{}c@{\hspace{2pt}}c@{\hspace{2pt}}c@{\hspace{2pt}}c@{\hspace{2pt}}c@{\hspace{2pt}}c@{}}
        \toprule
        \makecell[c]{\textbf{F15453}\\{{$\boldsymbol{k=1}$}}} & 
        \makecell[c]{\textbf{F33659}\\{$\boldsymbol{k=2}$}} & 
        \makecell[c]{\textbf{F65085}\\{$\boldsymbol{k=3}$}} & 
        \makecell[c]{\textbf{F2369}\\{$\boldsymbol{k=13}$}} & 
        \makecell[c]{\textbf{F58810}\\{$\boldsymbol{k=14}$}} & 
        \makecell[c]{\textbf{F21836}\\{$\boldsymbol{k=15}$}} \\
        \midrule
        translation & French & language & bienfaits & here & NameInMap \\
        Translation & France & Speaking & attentes & Here & CloseOperation \\
        translators & french & languages & prochaines & Below & Jspwriter \\
        \bottomrule
        \end{tabular}
        \end{footnotesize}
        }
        \label{tab:layer25_results_compact}
    \end{minipage}%
    \hfill
    \begin{minipage}{0.42\textwidth}
        \centering
        \vspace{0.25cm}
        \caption{Performance of instruction positions, including pre-instruction and post-instruction.}\label{tab:position-comparison}
        \setlength{\tabcolsep}{3pt} 
        \scalebox{1}{
        \begin{small}
        \begin{tabular}{cccc}
        \toprule
        Position & Strict Acc & Loose Acc & Original \\
        \midrule
        Pre-Instruction & 0.14 & 0.47 & 0.56 \\
        Post-Instruction & 0.23 & 0.64 & 0.75 \\
        \bottomrule
        \end{tabular}
        \end{small}}
    \end{minipage}
\end{table*}

\subsection{Analysis of Instruction-Related Concepts}\label{sec:analysis-concept}
To investigate RQ1, we analyze the interpretability of features extracted using SAEs and assess their correspondence to instruction-related concepts. Our analysis consists of two parts. First, we examine the activating text of extracted features with Neuronpedia~\cite{neuronpedia} to evaluate their semantic relevance to instructions. Second, we compare how strongly the activating examples of top-$k$ features and lower-ranked features correspond to instruction-related concepts, demonstrating the relationship between feature importance and instruction relevance.

 We focus on analyzing the consistent instruction-relevant latent activations through the lens of Neuronpedia~\cite{neuronpedia}, which provides detailed activated text for each SAE latent. Taking translation-related instructions as an example (e.g., ``Translate the sentence to French.''), we identify a notable latent that shows strong activation patterns. This latent exhibits high activation not only for various languages but also for directional prepositions like ``to'' and ``from'' that commonly appear in translation instructions, as shown in Table~\ref{tab:Feature15425}. We summarize two key findings as below:
\vspace{-3pt}
\begin{itemize}[leftmargin=*]\setlength\itemsep{-0.3em}
\item Our extracted SAE latent features show strong correspondence with instruction-related concepts, as demonstrated in Table~\ref{tab:Feature15425}. The extracted features consistently activate on instruction-relevant terms (e.g., ``translate'', ``French'') and related linguistic elements. 
\item The activating examples of our extracted top-$k$ features reveal a clear relevance pattern: they are directly corresponding to core instruction elements (e.g., task commands, target specifications), while those of lower-ranked features show decreasing relevance to instruction-relevant terms, capturing more peripheral or contextual information. The result is shown in Table~\ref{tab:layer25_results_compact}. Take the Layer 25 as an example, for the top-13th feature, the top 3 tokens are French words. But for the top-14th and 15th features, the top 3 tokens seem irrelevant to the instruction.
\end{itemize}


\begin{figure}[tb]
    \centering
    \includegraphics[width=\columnwidth]{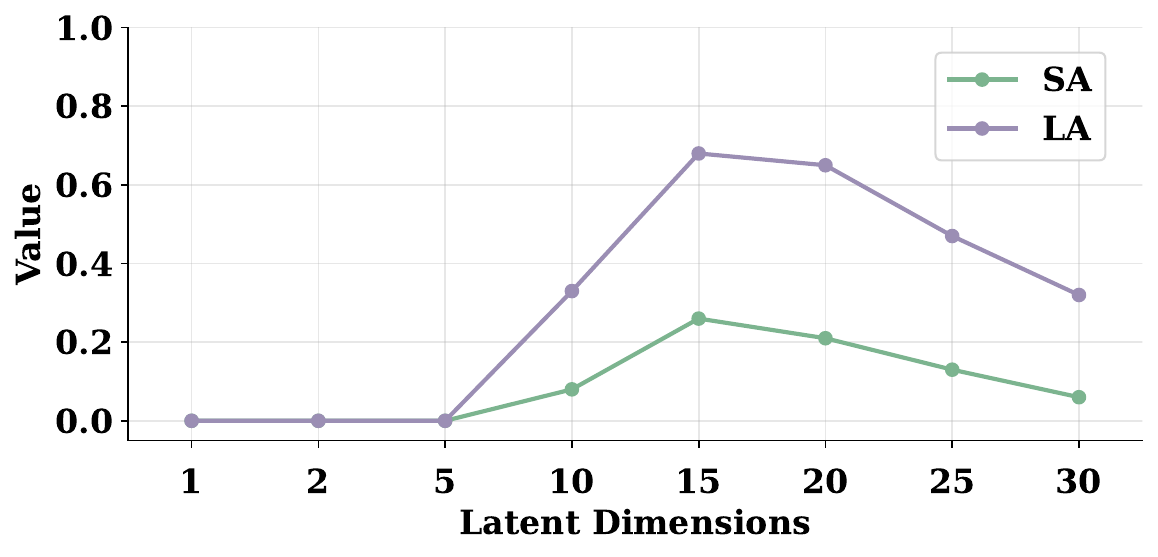}
    \caption{The impact of the number of latent dimensions (k) on our steering experiments. The x-axis represents different values of k, while the y-axis records the accuracy. We track the trend of strict accuracy (SA) and loose accuracy (LA) across 8 different k values. }
    \label{fig:latent-dimension-analysis}
\end{figure}

\begin{figure*}[tb]
    \centering
    \includegraphics[width=0.85\textwidth]{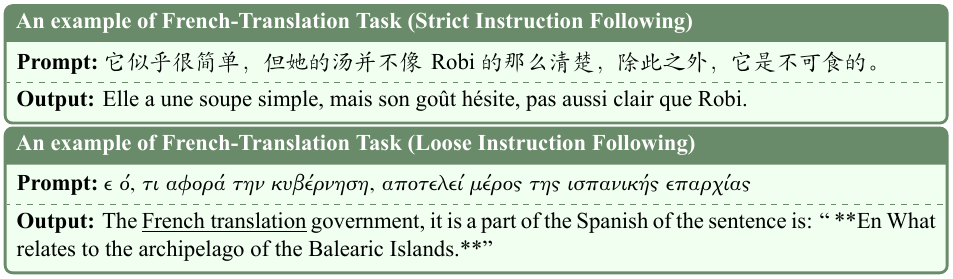}
    \caption{Examples of French translation task outcomes showing strict instruction following and loose instruction following using inputs in different languages. (Gemma-2-2b-it, SAE dimension of 65K)}
    \label{fig:instruction-examples}
\end{figure*}

\subsection{Steering Performance Analysis}
In this section, we evaluate the effectiveness of steering vectors constructed from SAE features and investigate the optimal number of features needed for reliable control. 

\begin{figure*}[tb]
    \centering
    \includegraphics[width=\textwidth]{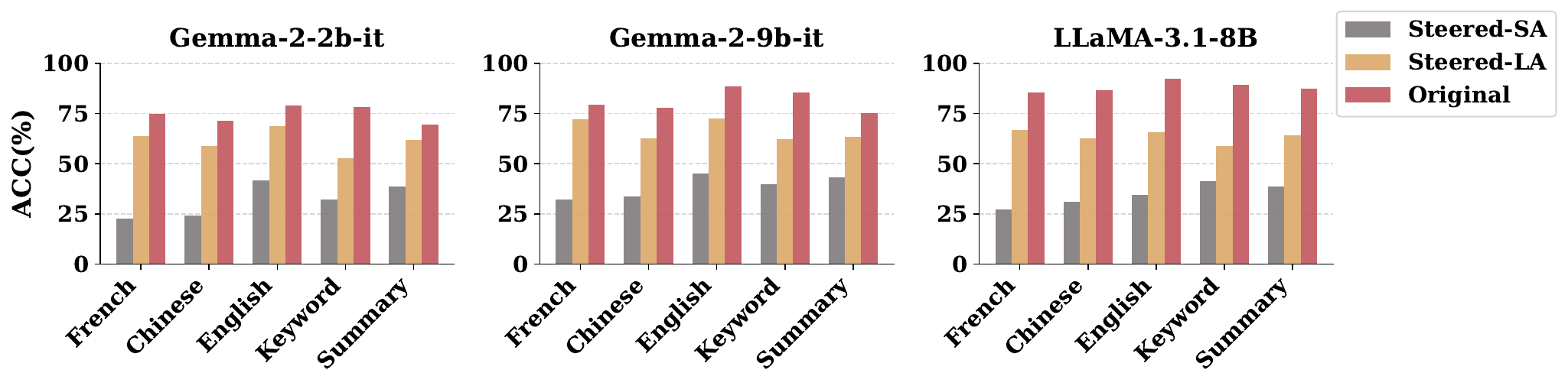}
    \caption{Performance comparison between original model outputs and two steering approaches across different instruction types on Gemma-2-2b-it and Gemma-2-9b-it models. Results show the accuracy percentages for translation tasks (French, Chinese, English), keyword inclusion, and summarization tasks. }
    \label{fig:steering-perforamnce-camparison}
\end{figure*}

\vspace{3pt}
\noindent\textbf{Steering Effectiveness.}\, 
We visualize a case study in Figure~\ref{fig:instruction-examples} and compare the performance of steering results in Figure~\ref{fig:steering-perforamnce-camparison}, including both \textit{strict accuracy} and \textit{loose accuracy}. Our analysis reveals several key findings:
\vspace{-3pt}
\begin{itemize}[leftmargin=*]\setlength\itemsep{-0.3em}
\item The quantitative results in Figure~\ref{fig:steering-perforamnce-camparison} demonstrate significant improvements in instruction following, with the steered models achieving over 30\% strict accuracy across different tasks. The loose accuracy of our steered approach performs nearly on par with prompting-based instruction methods, falling only slightly below. These results strongly indicate that SAIF can effectively extract features for user instructions and adjust LLMs' behaviors according to relevant instructions.
\item The case study in Figure~\ref{fig:instruction-examples} illustrates two distinct scenarios of instruction following: strict adherence (successful Chinese-to-French translation) and loose following (understanding that this is a French translation task). 
It demonstrates how SAIF manipulates model responses from the failure case toward either strict instruction following or loose instruction following.
\item The Gemma-2-9b-it model consistently outperforms Gemma-2-2b-it with slightly higher instruction steering performance across all five tasks, suggesting that SAIF's effectiveness scales well with model size.
\item The LLaMA-3.1-8B model shows comparable performance to the Gemma models across tasks. Looking at French translation as an example, LLaMA-3.1-8B achieves around 30\% strict accuracy and 65\% loose accuracy, which is similar to Gemma-2-2b-it's performance. 
\end{itemize}


\vspace{3pt}
\noindent\textbf{Latent Dimension Analysis.}\,
We study the effect of single latent and the number of latents on steering, showing that too few and too many dimensions both lead to failures. For individual latent, we use the single top 1 latent and latent listed in Table~\ref{tab:Feature15425} for steering. Despite their apparent semantic relevance to translation tasks, the model shows zero accuracy. This suggests that instruction following cannot be captured by a single high-level concept, even when that concept appears highly correlated with specific instruction types.

This observation leads us to investigate whether a combination of multiple latent dimensions could achieve better steering performance. Our experiments, shown in Figure~\ref{fig:latent-dimension-analysis}, systematically evaluate the impact of varying the number of latent dimensions from 1 to 30. The instructions used here are sourced from French translation task. The results reveal several key patterns:
\vspace{-3pt}
\begin{itemize}[leftmargin=*]\setlength\itemsep{-0.3em}
\item Steering performance remains near zero when $k \leq 5$, indicating that too few dimensions are insufficient for capturing instruction-following behavior. Performance begins to improve notably around $k=10$, with both strict accuracy and loose accuracy showing substantial increases.
\item The optimal performance is achieved at $k=15$, where loose accuracy peaks at approximately $0.7$ and strict accuracy reaches about $0.25$.
\item However, as we increase dimensions beyond $k=15$, both metrics show a consistent decline. This deterioration becomes more pronounced as $k$ approaches $30$, suggesting that excessive dimensions introduce noise that interferes with effective steering.
\end{itemize}

\begin{table*}[htbp]
    \centering
    \small
    \caption{Analysis of Layer Features}
    \begin{tabular}{@{\hspace{0.5cm}}cccc@{\hspace{0.5cm}}}
    \toprule
    \textbf{\# of Layer} & \textbf{\parbox[c][0.7cm][c]{7cm}{\centering Top 5 tokens with the highest logit increases by the feature influence}} & \textbf{\# of top\_k} & \textbf{\# of Feature} \\
    \midrule
    25 & French, France, french, FRENCH, Paris & 2 & 33659 \\
    24 & French, nb, french, Erreur, Fonction & 8 & 65238 \\
    23 & French, France, french, Paris, Francis & 15 & 49043 \\
    22 & English, english, Spanish, French, Hindi & 12 & 351 \\
    21 & Belgian, Belgium, Brussels, Flemish, Belgique & 14 & 27665 \\
    \bottomrule
    \end{tabular}
    \label{tab:layerwise_analysis}
\end{table*}

\subsection{The Role of Last Layer Representations in Instruction Processing}
In previous sections, we exclusively used SAE from the last Transformer layer for concept vector extraction and instruction steering. In this section, we analyze why extracting concepts and steering from the final layer is most effective.

\vspace{3pt}
\noindent\textbf{Concept Extraction Perspective.}\, 
From the results in Table~\ref{tab:layerwise_analysis}, we observe an intriguing phenomenon that shallower layers are less effective in providing clean instruction-relevant features. Following our default experimental settings, we extract the top 15 SAE features from each layer of the model. The features extracted from the last layer can precisely capture the semantics of `French', showing strong activations on French-related words, where $k=2$ indicates this feature is considered the second most instruction-relevant feature. Starting from the penultimate layer, as we attempt to trace French-related features, our experimental results reveal that the extracted French-related concepts undergo a gradual shift as the layer depth decreases. Specifically, the feature evolves from exclusively activating on French-related tokens to encompassing a broader spectrum of languages (English, Spanish, Hindi, and Belgian), demonstrating a hierarchical abstraction pattern from language-specific to cross-lingual representations. Moreover, the increasing $k$ values suggest that these French-related features become less instruction-relevant in earlier layers. For Gemma2-2b-it model, before Layer 21, we can no longer identify French-related features among the top 15 SAE features.

\vspace{3pt}
\noindent\textbf{Steering Perspective.}\, 
We conducted steering experiments using the top 15 features extracted from Layers 21-25 respectively under default settings on French Translation task. The results align with our findings on concept extraction, showing the effectiveness and importance of last layer representation on instruction following. Using loose accuracy as the evaluation metric, we observe that steering with Layer 24 features still maintains some effectiveness, though the loose accuracy drops sharply from 0.64 (Layer 25) to 0.33. Steering attempts using features from earlier layers fail to guide the model towards instruction-following behavior, with the model instead tending to generate repetitive and instruction-irrelevant content.

\subsection{Impact of Instruction Position}\label{sec:4.5}
Previous studies have shown that models' instruction-following capabilities can vary significantly depending on the relative positioning of instructions and content. This motivates us to examine how instruction positioning affects the activation patterns of previously identified features.

We investigate the effect of instruction position by comparing two patterns: pre-instruction ($P_{pre}$ = [Instruction] + [Content]) and post-instruction ($P_{post}$ = [Content] + [Instruction]) as in~\citet{liu2023instruction}. Using identical instruction-content pairs while varying only their relative positions allows us to isolate the effects of position. Our analysis reveals several key findings from both the quantitative metrics (see Table~\ref{tab:position-comparison}) and feature activation patterns (see Figure~\ref{fig:pre-post-instruction}):
\vspace{-5pt}
\begin{itemize}[leftmargin=*]\setlength\itemsep{-0.3em}
\item Performance metrics demonstrate that post-instruction positioning consistently outperforms pre-instruction, with post-instruction achieving higher accuracy across all measures (Strict Acc: 0.23 vs 0.14, Loose Acc: 0.64 vs 0.47), aligning with the result in~\citet{liu2023instruction}.
\item Feature activation patterns show that post-instruction enables more robust processing with stronger activation peaks (particularly for key features like F33659), more consistent stability scores, and higher activation probabilities (>80\%) across most features compared to pre-instruction's more variable patterns.
\end{itemize}

\section{Conclusions}
In this paper, we have introduced to use SAEs to analyze instruction following in LLMs, revealing the underlying mechanisms through which models encode and process instructions. Our analysis demonstrates that instruction following is mediated by interpretable latent features in the model's representation space
We have developed a lightweight steering technique that enhances instruction following by making targeted modifications to specific latent dimensions. We find that effective steering requires the careful combination of multiple latent features with precisely calibrated weights. 
Extensive experiments across diverse instruction types have demonstrated that our proposed steering approach enables precise control over model behavior while consistently maintaining coherent outputs.

\clearpage
\section*{Limitations}
One limitation of our steering approach is that it sometimes produces outputs that only partially follow the intended instructions, particularly when handling complex tasks. While the model may understand the general intent of the instruction, the generated outputs may not fully satisfy all aspects of the requested task. For example, in translation tasks, the model might incorporate some elements of the target language but fail to produce a complete and accurate translation. Besides, our current work focuses primarily on simple, single-task instructions like translation or summarization. In future, we plan to investigate how to extend this approach to handle more sophisticated instruction types, such as multi-step reasoning tasks or instructions that combine multiple objectives. Additionally, our experiments were conducted using models from the Gemma and Llama two LLM families. In the future, we plan to extend this analysis to a more diverse set of language model architectures and families to validate the generality of our findings.

\bibliography{ref}

\clearpage
\appendix

\onecolumn

\noindent

\section{Details of Instructions}
\label{app:a}
\begin{table*}[!h]
\centering
\includegraphics[width=0.95\textwidth]{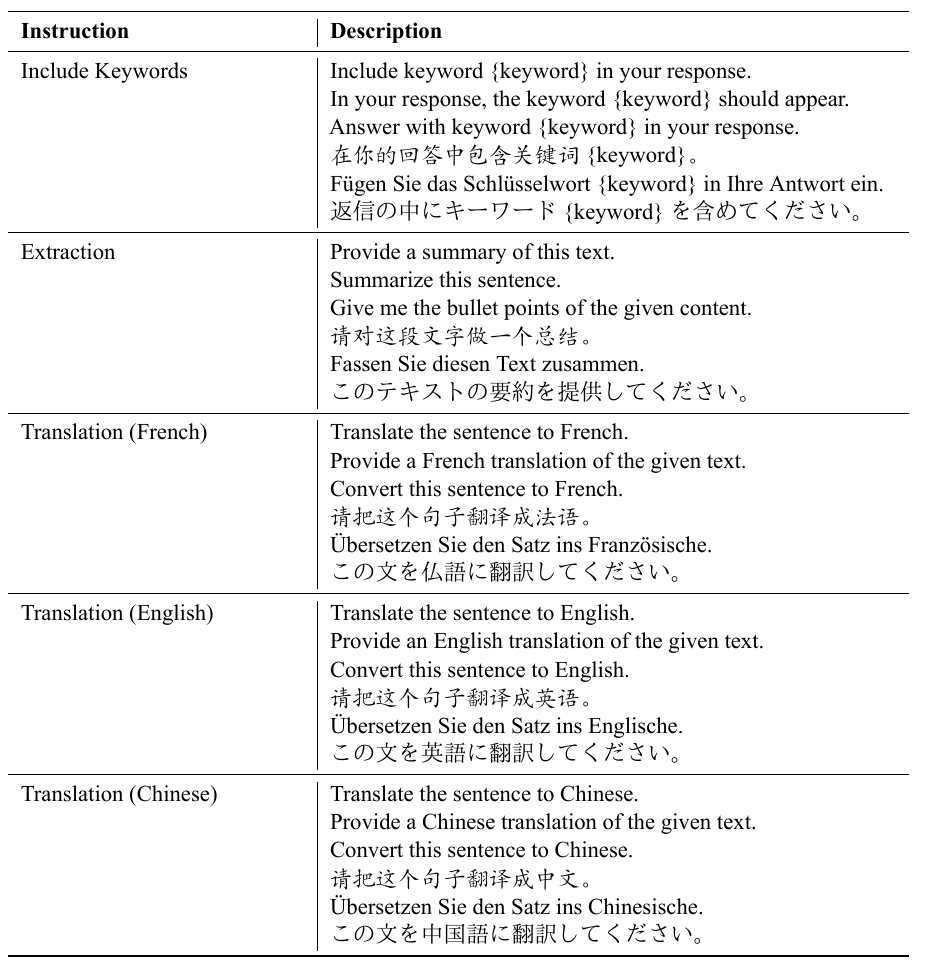}
\end{table*}

\section{Related Work}
In this section, we briefly summarize several research directions that are most relevant to ours.

\paragraph{Instruction Following in Language Models.}
Instruction following capabilities are crucial for improving LLM performance and ensuring safe deployment. Recent advances in instruction tuning have demonstrated significant progress through various methods \cite{ouyang2022training,sanh2022multitask,wei2022finetuned,chung2024scaling}. However, capable models still struggle with hard-constrained tasks \cite{sun2023evaluating} and lengthy generations\cite{li2024measuring}. Some studies find that instruction following can be improved with in-context few-shot examples~\cite{kung2023models}, optimal instruction positions~\cite{liu2023instruction}, carefully selected instruction-response pairs with fine-tuning~\cite{zhou2024lima}, and adaptations~\cite{hewitt2024instruction}. Unfortunately, the mechanistic understanding of how LLMs internally represent and process these instructions remains limited. 

\paragraph{Language Model Representations.}
A body of research have focused on studying the linear representation of concepts in representation space~\cite{kim2018interpretability}. The basic idea is to find a direction in the space to represent the related concept. This can be achieved using a dataset with positive and negative samples relevant to concepts. Existing approaches computing the concept vectors include probing classifiers~\cite{belinkov2022probing}, mean difference~\cite{rimsky2024steering,zou2023representation}, mean centering~\cite{jorgensen2024improving}, gaussian concept subspace~\cite{zhao2025beyond}, which provide a rich set of tools to derive concept vectors. The derived concept vectors represent various high-level concepts such as honesty~\cite{li2024inference}, truthfulness~\cite{tigges2023linear}, harmfulness~\cite{zou2023representation}, and sentiments~\cite{zhao2025beyond}.

\paragraph{Sparse Autoencoders.} Dictionary learning is effective in disentangling features in superposition without representation space. Sparse autoencoder (SAE) offers a feasible way to map representations into a higher-dimension space and reconstruct to representation space. Various SAEs have been proposed to improve their performance such as vallina SAEs~\cite{sharkey2022sae}, TopK SAEs~\cite{gao_scaling_2024}. Based on them, a range of sparse autoencoders (SAEs) have been trained to interpret hidden representations including Gemma Scope~\cite{lieberum2024gemma} and Llama Scope~\cite{he2024llama}. These SAEs have also been used to interpret models' representational output~\cite{kissane2024interpreting} and understand their abilities~\cite{ferrando2025do}.

\paragraph{Activation Steering.}
Recently, a body of research has utilized concept vectors to steer model behaviors during inference. Specifically, concepts vectors can be computed with diverse approaches, and these vectors are mostly effective on manipulating models generating concept-relevant text. For instance, many studies find it useful in improving truthfulness\cite{marks2023geometry} and safety~\cite{arditi2024refusal}, mitigating sycophantic and biases~\cite{zou2023representation}. Steering primarily operates in the residual stream following methods defined in Eq.~\eqref{eq:steer}, but it is worth-noting that the steering vectors can be computed from either residual stream representations or SAEs. Existing work mostly concentrates on computing with residual stream representations, which provide limited insights on what finer features contribute to the high-level concept vector. This coarse approach could further limit our deeper understanding on more complicated vectors such as instructions. In our work, we aim to bridge this gap by studying instruction vectors with SAEs to uncover their working mechanism.

\begin{figure}[tb]
    \centering
    \includegraphics[width=0.6\columnwidth]{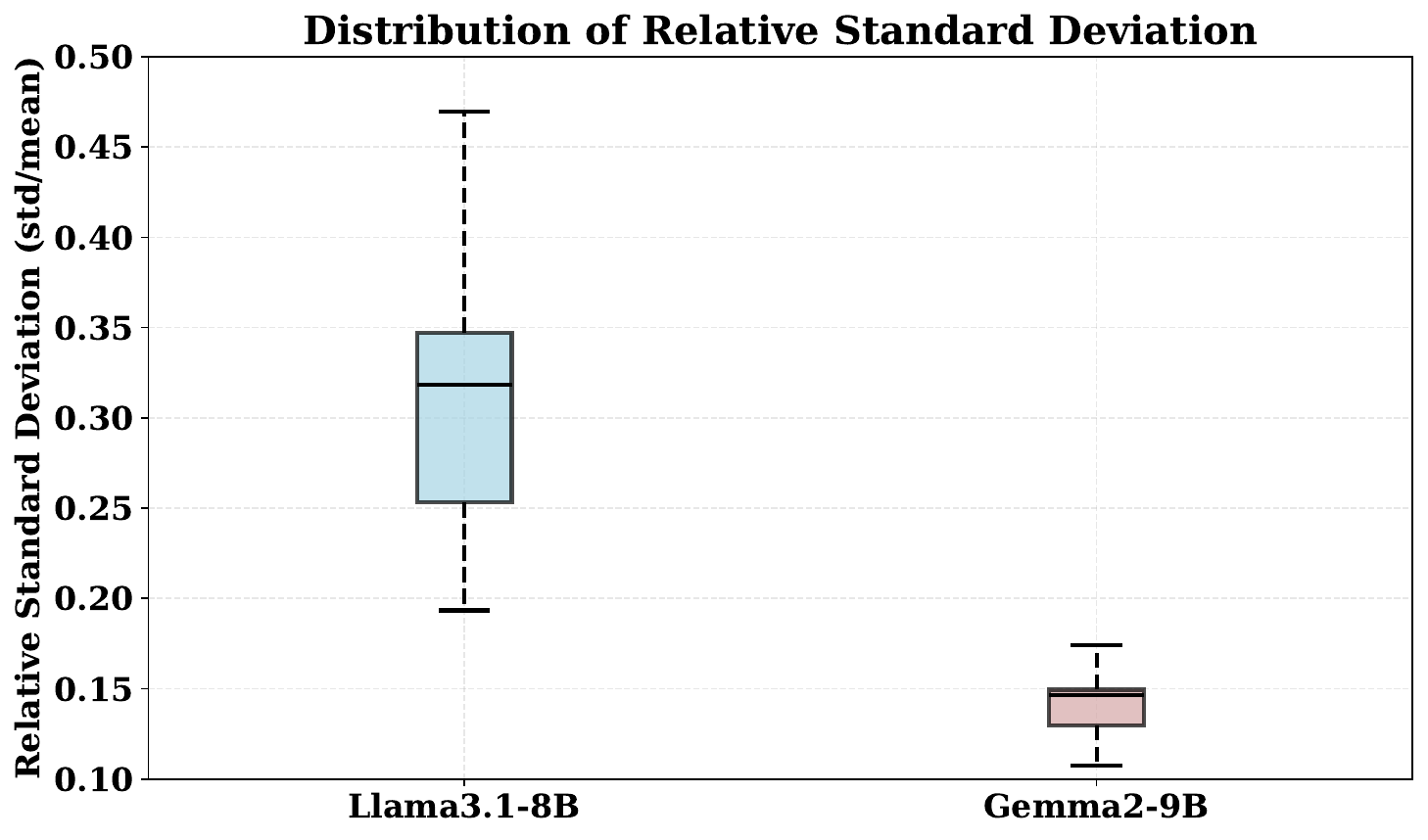}
    \caption{Visualization of steering vectors extracted from LLaMA-3.1-8B and Gemma-2-9B for French translation task. The y-axis denotes the ratio between the standard deviation and mean of feature activation strengths.}
    \label{fig:cross-model}
\end{figure}

\section{Additional Results for Llama-3.1-8b}\label{app:C}
In our experimental setup, we employ Equation~(\ref{equ:st}) to control feature activation during model steering, where $\mu_i$ denotes the pre-computed mean activation strength and $s_i$ represents the standard deviation for feature $i$. The hyperparameter $\beta$ controls the perturbation magnitude relative to the standard deviation.
Our experiments reveal distinct robustness characteristics across different model architectures. For the Gemma-2 family models, the steering vectors maintain their effectiveness when $\beta \in [-1,1]$, indicating robust feature representations. These models exhibit high activation strength values ($\mu_i$) with low standard deviations ($s_i$), suggesting stable and consistent feature characteristics.
In contrast, the Llama-3.1-8b model demonstrates higher sensitivity to activation perturbations. The steering vectors remain effective only when $\beta \in [-0.1,0.1]$, indicating a significantly narrower tolerance range. The relative standard deviations illustrated in Figure~\ref{fig:cross-model} quantify this distinction. This narrow tolerance range suggests that Llama-3.1-8b's feature space may possess the following characteristics: stricter boundaries between features, more discrete transitions between different instruction states, and poorer robustness to noise.

\section{Steering Accuracy Evaluation based on GPT-4o-mini}\label{sec:gpt-4o-evaluation-details}
To evaluate generated outputs, we instruct GPT-4o-mini to rate in the following way.
For each instance, we provide GPT-4o-mini with three components: the original input text, the instruction, and the model-generated output. To ensure reliable assessment, we implement a voting mechanism where GPT-4o-mini performs five independent evaluations for each instance. For each evaluation, GPT-4o-mini is prompted to assess the instruction following level by selecting whether the generated content completely follows the instruction (A), contains instruction keywords but doesn't follow the instruction (B), or is completely irrelevant to the instruction (C). The final grade is determined by majority voting among the five evaluations. In cases where there is no clear majority (e.g., when votes are split as 2-2-1), we choose the lower grade between the two options that received the most votes (C is considered lower than B, and B is lower than A). This ensures a stringent evaluation standard when the votes are divided. Thus, the \textit{Strict Accuracy} is the ratio of A and the \textit{Loose Accuracy} is the ratio of A + B. The prompt we use in the experiments can be found in Table~\ref{tab:evaluation}.

\begin{table*}[tbp]
\centering
\caption{Evaluation Prompt for Generated Output}
{\ttfamily\small
\begin{tabular}{p{\textwidth}}
\toprule
Your task is to strictly evaluate whether the generated output follows the given instruction.\\
First you should review the following components:\\[0.5em]
Original Input: \{input\_text\}\\
Instruction: \{instruction\}\\
Generated Output: \{generated\_output\}\\[0.5em]
Here is the evaluation criteria:\\[0.5em]
A: The generated content completely follows the instruction.\\
B: Contains instruction keywords but doesn't follow the instruction completely.\\
C: Completely irrelevant to the instruction Critical.\\[0.5em]
Remember:\\[0.5em]
If the Generated Output only contains repeated words or sentences, select C immediately. \\[0.5em]
DO NOT provide explanation. Provide your evaluation by selecting one option(A/B/C).\\
Your Answer is: \\
\bottomrule
\end{tabular}
\label{tab:evaluation}
}
\end{table*}

\section{Model Scale Analysis}
We explore the influence of both model scale and SAE scale, showing larger sizes always contribute to better performance. Using SAE with larger dimensions (e.g., increasing Gemma-2-2b's SAE from 16K to 65K) can effectively improve the interpretability of feature extraction. For the same prompt, Gemma-2-2b's 16K SAE is almost unable to extract interpretable features under our settings, while the 65K model performs well. For Gemma-2-9b and Llama3.1-8b models, even the SAE with minimal dimensions can extract features with good interpretability. 

\clearpage
\section{More Activating Examples of Top-ranked Features }

\begin{longtable}{@{}>{\centering\arraybackslash\small}p{0.9\textwidth}@{}}
\caption{The remaining eight features we used to construct the steering vector for Gemma2-2B SAE on the French Translation task, along with their corresponding activation examples. (The other seven features can be found in Table~\ref{tab:Feature15425} and Table~\ref{tab:layer25_results_compact}.) The examples are provided by Neuropedia (\textcolor{green}{Lin \& Bloom}, \textcolor{green}{2024}).}\\
\toprule
{Layer25, Feature42374} \\
\midrule
Could you please translate the following \deephl{sentence} to French? \\
I \lighthl{think} ``everyone'' \lighthl{and} ``we'' are the same in \lighthl{this} \deephl{sentence}. \\
\bottomrule\\
\toprule
{Layer25, Feature49454} \\
\midrule
Quote from the \lighthl{article} \deephl{below} \lighthl{:} Variable names are case - sensitive. \\
With pure \lighthl{mind} and internal \deephl{comtemplation} \deephl{there} is no \lighthl{need} for... \\
\bottomrule\\
\toprule
{Layer25, Feature54902} \\
\midrule
The incredible spe ta culo \lighthl{de} \lighthl{la} \deephl{vida}, the \lighthl{incredible} spe ta \deephl{culo} \deephl{de} \deephl{la} \deephl{muerte}! \\
This is a continuation of the precedent the band established \deephl{with} Re... \\
\bottomrule\\

\toprule
{Layer25, Feature55427} \\
\midrule
Whatever the \lighthl{modifier} may \lighthl{be}, both \deephl{sentences} \deephl{are} discussing... \\
I can \deephl{make} no distinction \lighthl{between} \lighthl{the} \lighthl{two} \lighthl{}l{sentences} at issue...\\
\bottomrule\\

\toprule
{Layer25, Feature6201} \\
\midrule
Furthermore, figure has a plethora of other \lighthl{senses}, evinced \lighthl{by} \lighthl{the}
\deephl{dictionary} \lighthl{entry} \lighthl{linked} above.\\
The \lighthl{meaning} and nuance of this \lighthl{phrase}
can be quite different depending on the \lighthl{context}.\\
\bottomrule\\

\toprule
{Layer25, Feature17780} \\
\midrule
How to convert the text into Hyperlinks? \deephl{Thanks} in \lighthl{advance}!\\
\deephl{Hi} Jimmy, I don't have your grandfather Birl listed in my files...\\
\bottomrule\\

\toprule
{Layer25, Feature22091} \\
\midrule
She can't focus sufficiently to \lighthl{utter} \deephl{complete} \lighthl{sentences} without needing to \lighthl{stop} and reflect.\\
He speaks in a Hiroshima accent and often \deephl{ends} his \lighthl{sentences} \lighthl{with} "~garu" and "~ja".\\
\bottomrule\\

\toprule
{Layer25, Feature59061} \\
\midrule
Helderberg is a \lighthl{Dutch} \deephl{name} \lighthl{meaning} "clear mountrain". \\
Kaila - Altered form \lighthl{of} \lighthl{English} Kaylay, \lighthl{meaning} "slender".\\
\bottomrule

\end{longtable}

\clearpage
\section{Examples of Instruction Following Tasks with Steering Vectors}\label{app:instruct-example}
\begin{center}
    \includegraphics[width=0.93\textwidth]{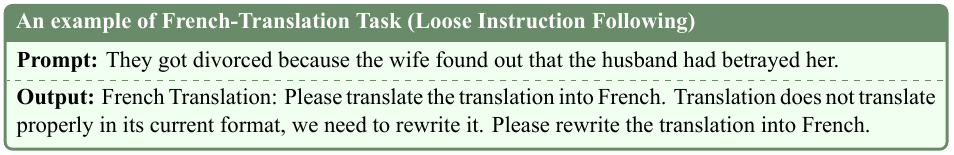}\\\vfill
    \includegraphics[width=0.93\textwidth]{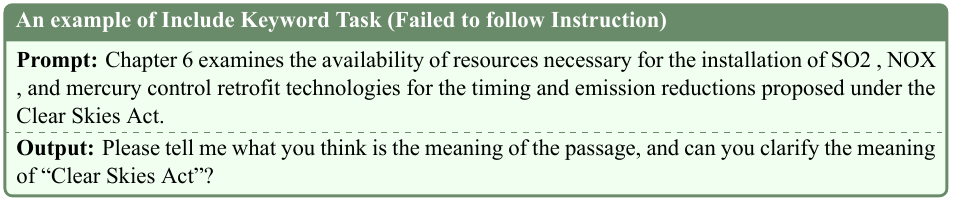}\\\vfill
    \includegraphics[width=0.93\textwidth]{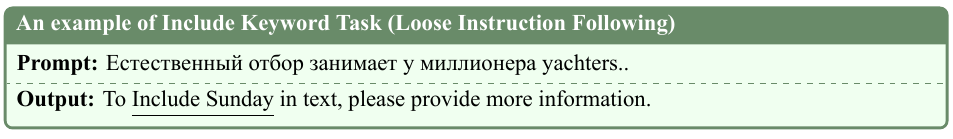}\\\vfill
    \includegraphics[width=0.93\textwidth]{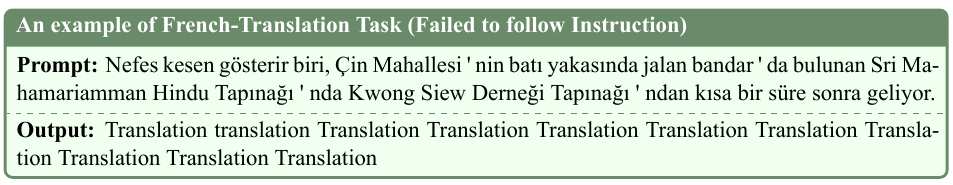}\\\vfill
    \includegraphics[width=0.93\textwidth]{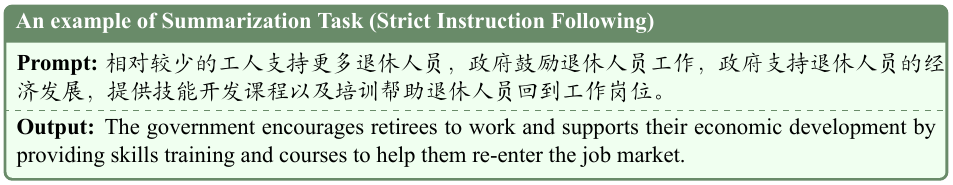}\\\vfill
    \includegraphics[width=0.93\textwidth]{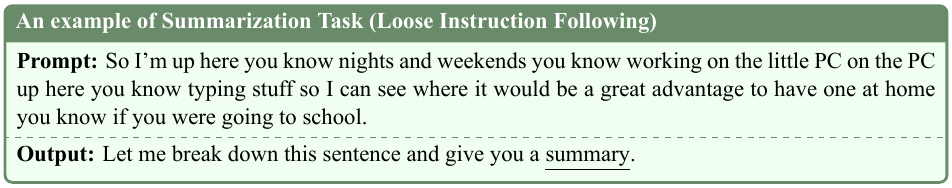}\\\vfill
    \includegraphics[width=0.93\textwidth]{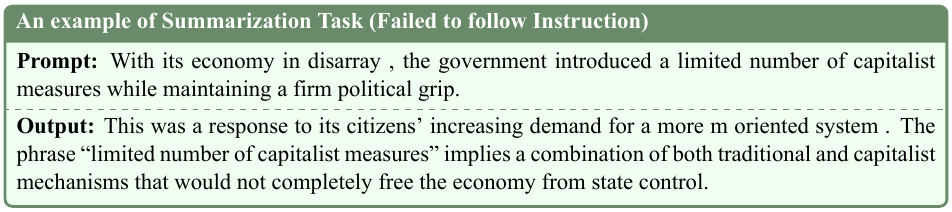}\\\vfill
\end{center}

\clearpage
\section{Extracted Features Correlation Visualization and Analysis}
\begin{figure*}[!h]
    \centering
    \includegraphics[width=\textwidth]{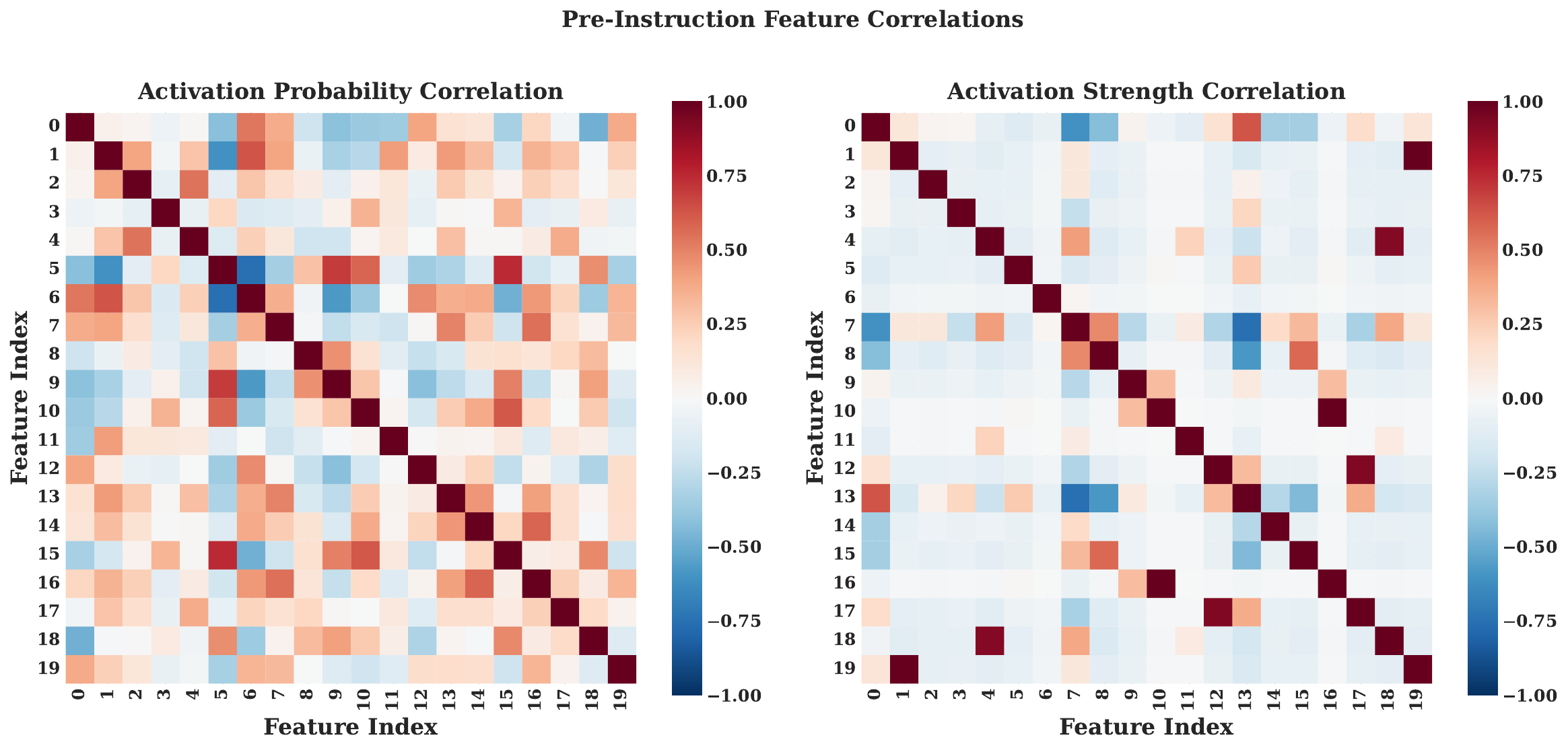}\\
    \includegraphics[width=\textwidth]{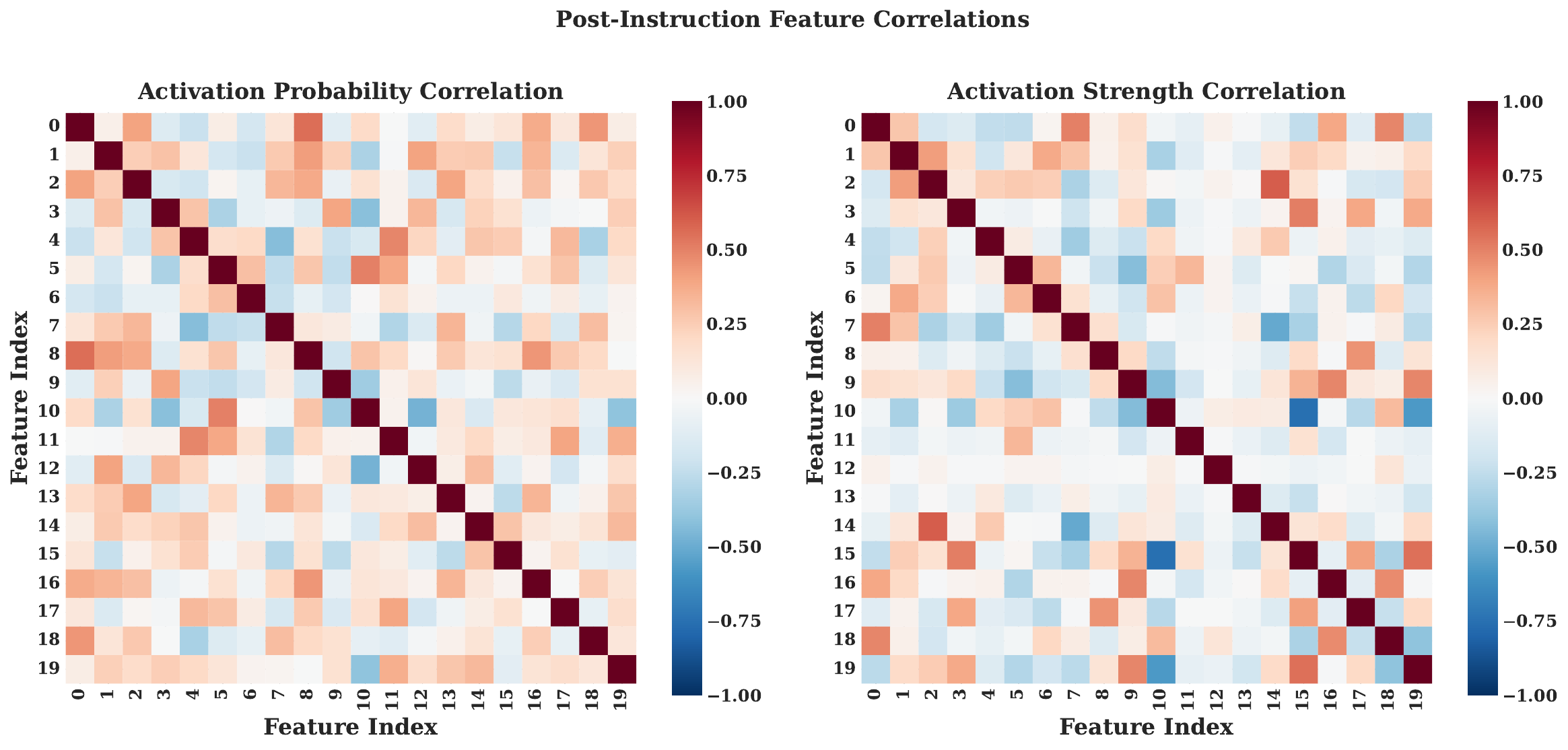}
    \caption{Heatmaps for Keyword Task.}
    \label{fig:7}
\end{figure*}

\begin{figure*}[!h]
    \centering
    \includegraphics[width=\textwidth]{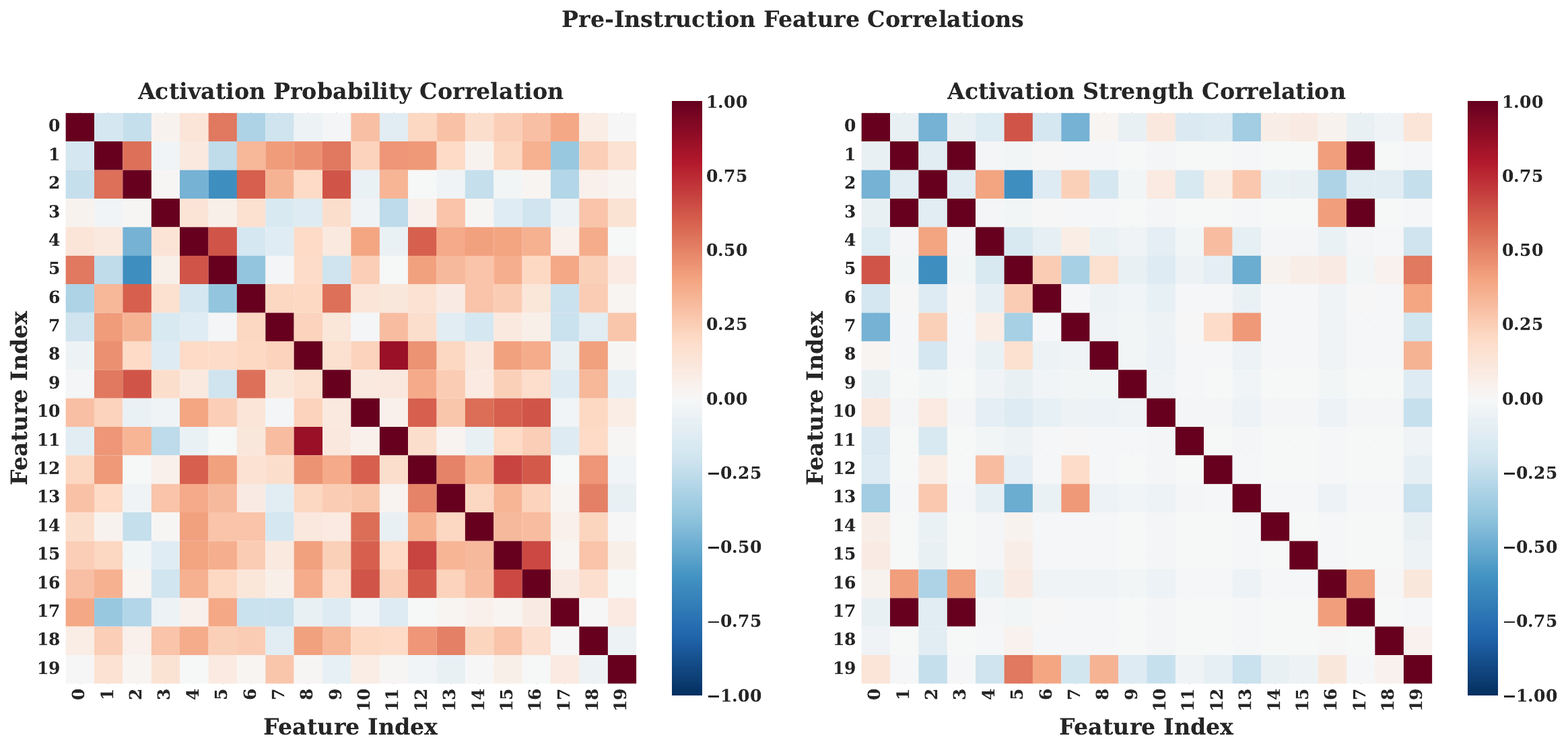}
    \includegraphics[width=\textwidth]{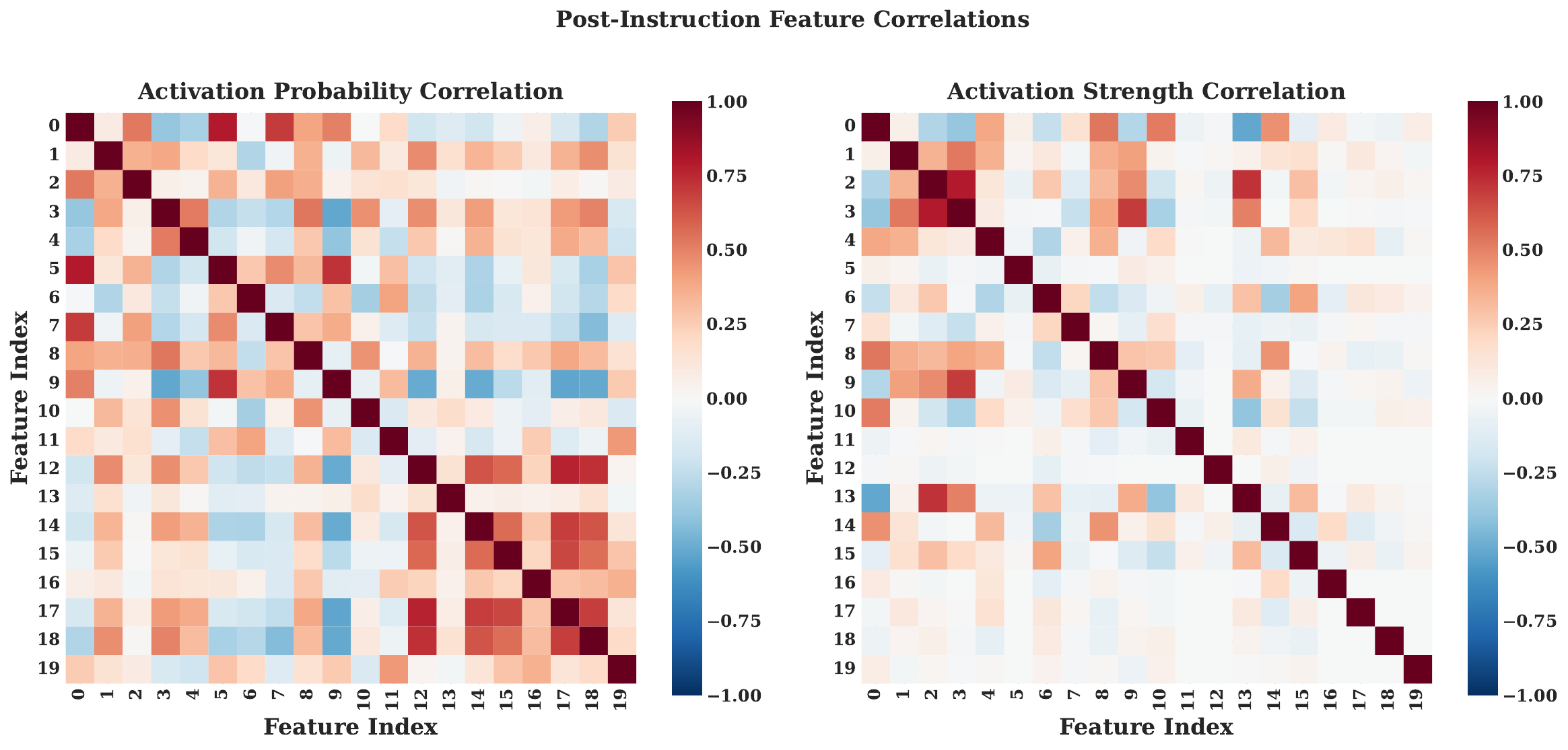}
    \caption{Heatmaps for Translation(Chinese) Task.}
    \label{fig:11}
\end{figure*}

In Section~\ref{sec:4.5}, we explored how instruction placement (before or after the original prompt) affects model behavior. To further understand how the model encodes and processes instructions in different positions, we present visualization analysis using feature correlation heatmaps.
Figure~\ref{fig:7} to Figure~\ref{fig:11} show the feature correlations of Gemma-2-2b model across five different tasks. 

Taking Figure~\ref{fig:7} as an example, the visualization is divided into Pre-Instruction and Post-Instruction modes. Each part contains two 20×20 heatmap matrices showing Activation Probability and Activation Strength correlations respectively. The heatmaps use a red-blue color scheme, where dark red indicates strong positive correlation (1.0), dark blue indicates strong negative correlation (-1.0), and light or white areas indicate correlations close to 0. The axes range from 0 to 19, representing the top 20 SAE latent features.

Our analysis reveals distinct differences between the two instruction placement modes. The Pre-Instruction mode shows dispersed correlations with predominantly light colors outside the diagonal, indicating stronger feature independence. In contrast, the Post-Instruction mode exhibits more pronounced red and blue areas, demonstrating enhanced feature correlations and a more tightly connected feature network.
This finding aligns with our key conclusion that effective instruction following requires precise combinations of multiple latent features. The stronger feature correlations in Post-Instruction mode confirm that single-feature manipulation is insufficient for reliable control. This insight into feature cooperation supports the effectiveness of our proposed steering technique based on precisely calibrated weights across multiple features.

\begin{figure*}[!h]
    \centering
    \includegraphics[width=0.95\textwidth]{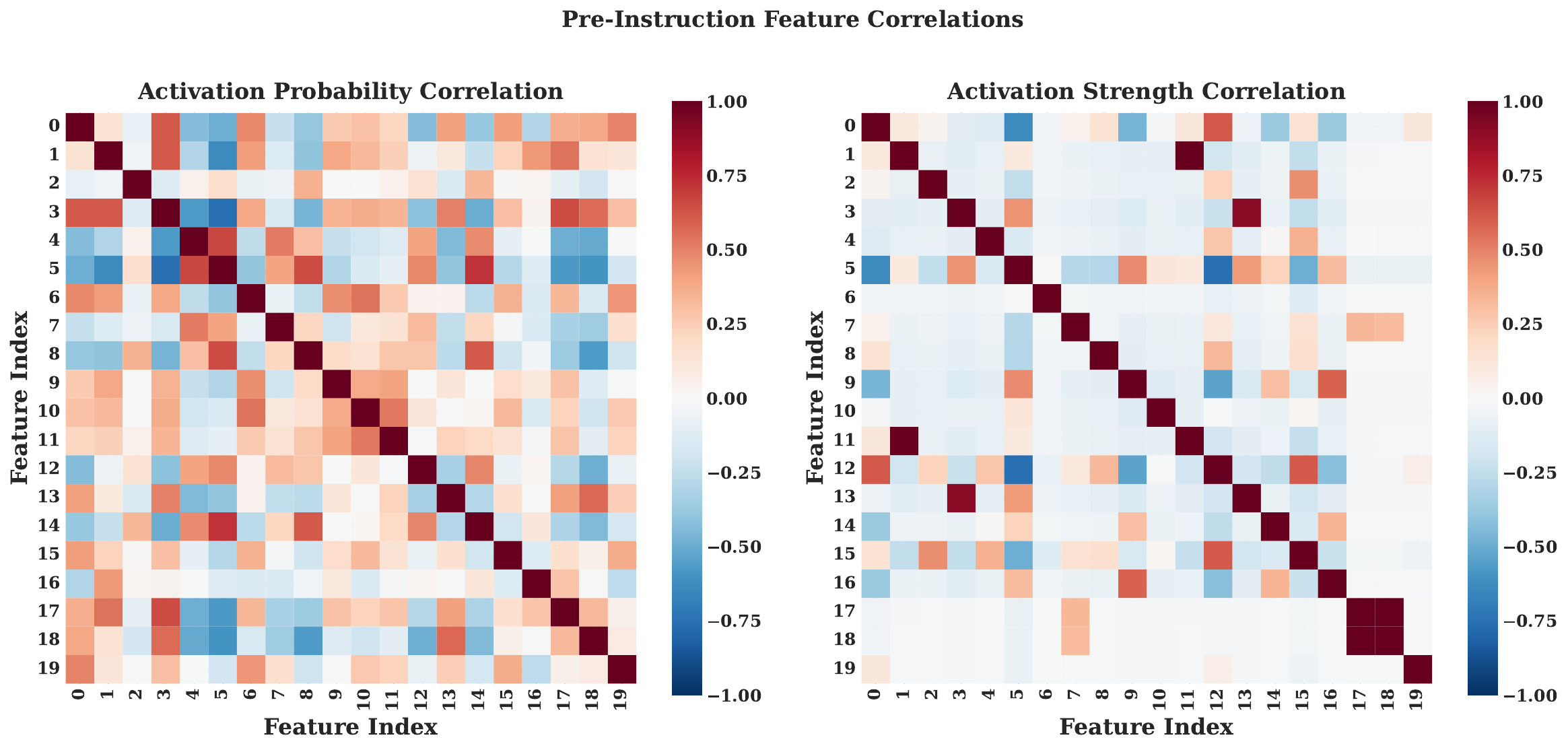}\\
    \includegraphics[width=\textwidth]{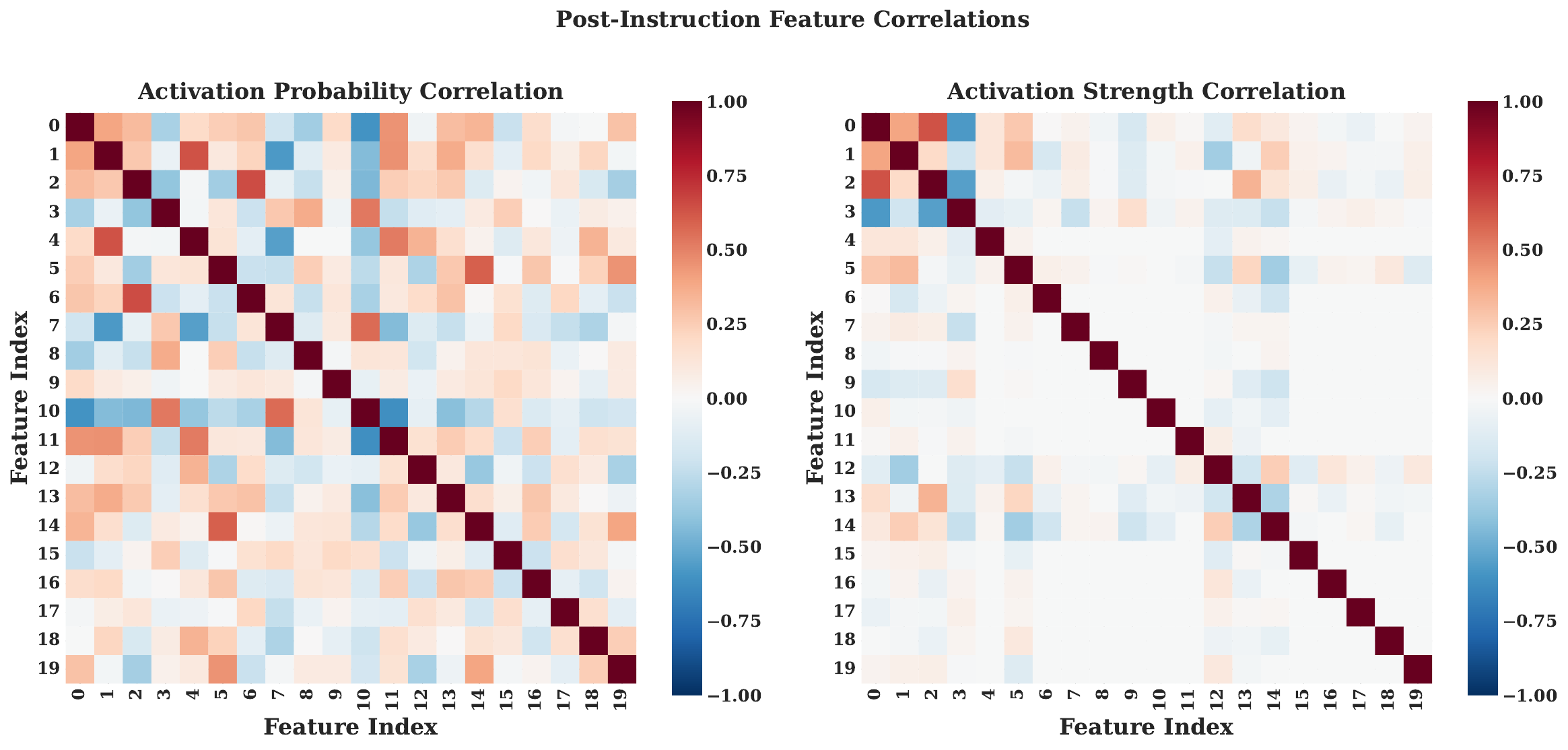}
    \caption{Heatmaps for Summarization Task.}
\end{figure*}

\begin{figure*}[!h]
    \centering
    \includegraphics[width=\textwidth]{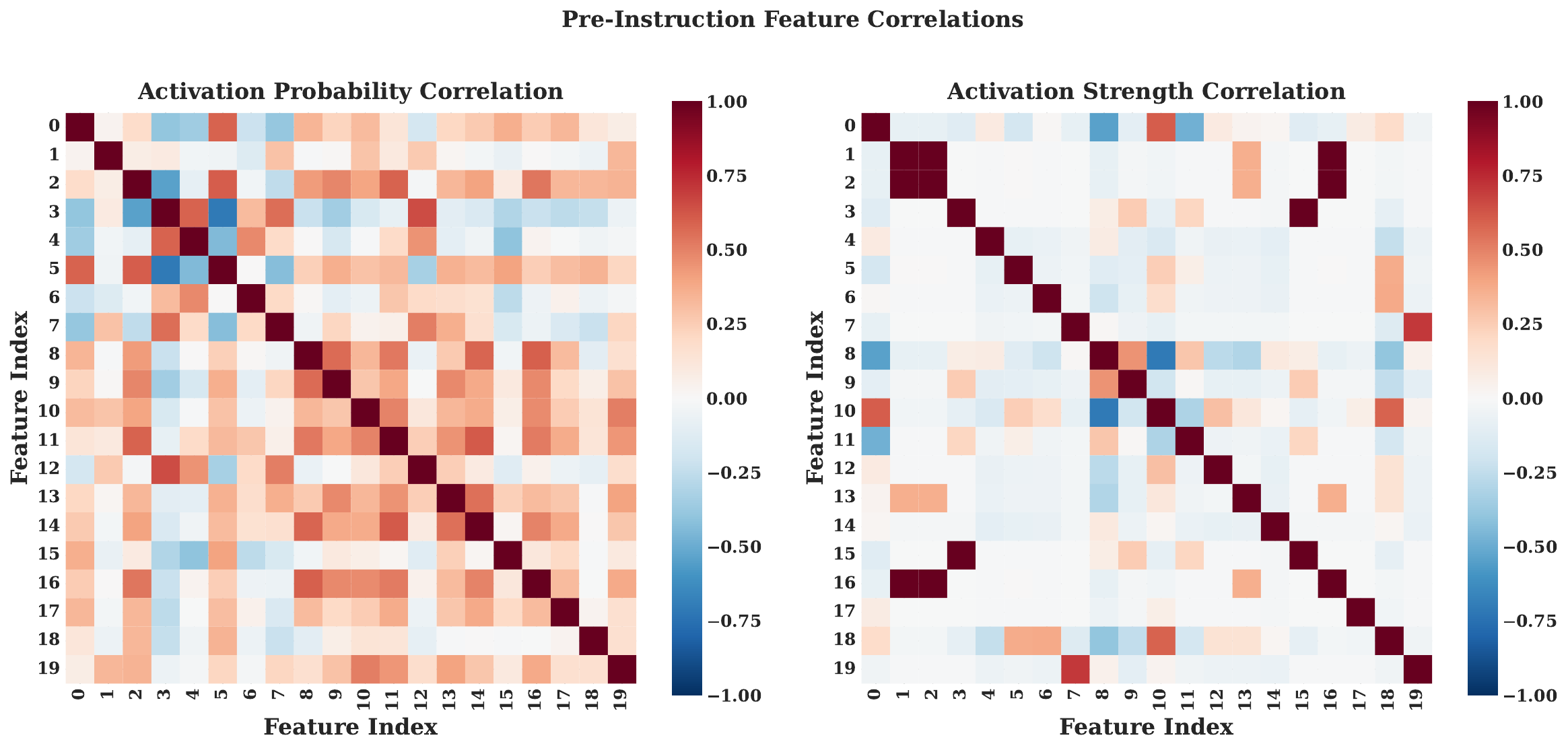}
    \includegraphics[width=\textwidth]{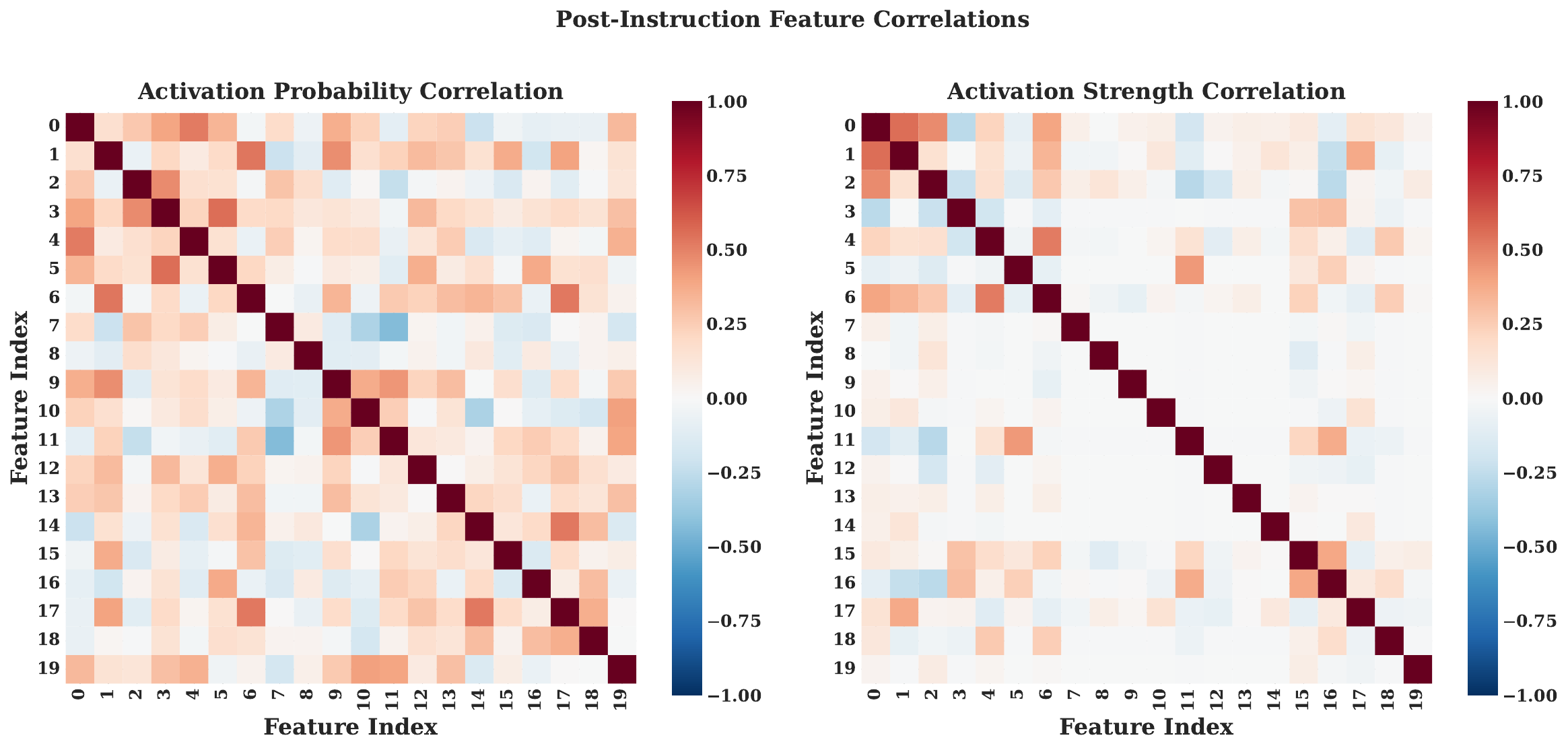}
    \caption{Heatmaps for Translation(English) Task.}
\end{figure*}

\begin{figure*}[!h]
    \centering
    \includegraphics[width=\textwidth]{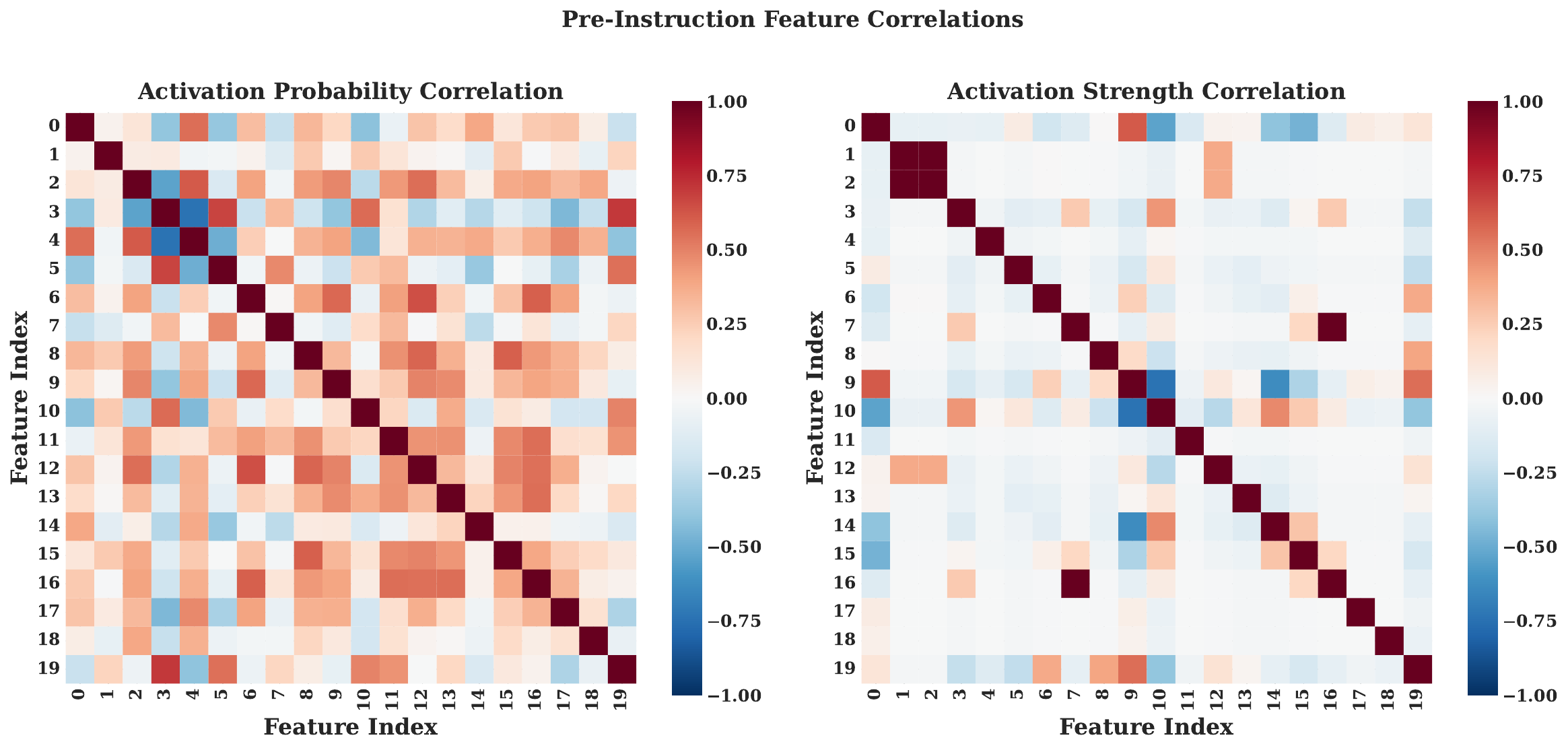}
    \includegraphics[width=\textwidth]{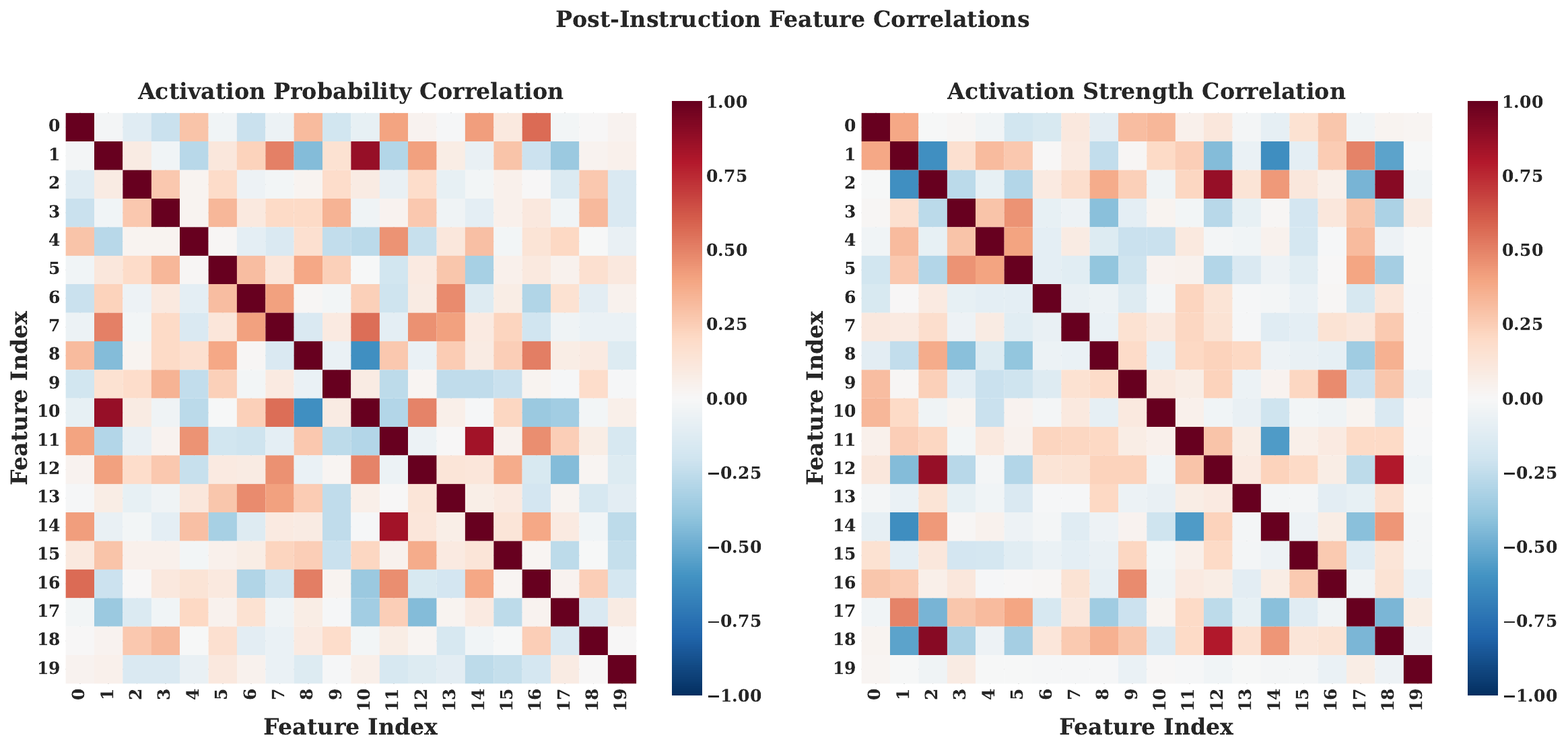}
    \caption{Heatmaps for Translation(French) Task.}
\end{figure*}

\end{document}